\documentclass[10pt,journal,compsoc]{IEEEtran}
\usepackage{graphicx}
\ifCLASSOPTIONcompsoc
  \usepackage[nocompress]{cite}
\else
  \usepackage{cite}
\fi

\usepackage[utf8]{inputenc}
\usepackage{color}
\usepackage{xcolor}
\usepackage{amssymb}
\usepackage{amsmath}
\usepackage{enumerate}
\usepackage{algorithm}
\usepackage[noend]{algpseudocode}
\usepackage{tabularx}
\usepackage{mathtools}
\usepackage{graphicx}

\usepackage[caption=false]{subfig}
\usepackage{mathrsfs}
\usepackage{float}
\usepackage{placeins}
\usepackage{array}
\newcolumntype{C}[1]{>{\centering\arraybackslash}m{#1}}
\usepackage{booktabs,adjustbox}
\usepackage[mathcal]{eucal}

\usepackage{leftidx}

    \newtheorem{theorem}{Theorem}[section]
    
    \newtheorem{definition}[theorem]{Definition}

    \newcommand{\qed}{\nobreak \ifvmode \relax \else
          \ifdim\lastskip<1.5em \hskip-\lastskip
          \hskip1.5em plus0em minus0.5em \fi \nobreak
          \vrule height0.75em width0.5em depth0.25em\fi}

\graphicspath{{img/}}

\newcommand\DTW{\text{DTW}}

\newcommand{\argmin}{\arg\!\min}

\begin{document}

\title{Times series averaging and denoising from a probabilistic perspective on time-elastic kernels 
}


\author{Pierre-Francois Marteau,~\IEEEmembership{Member,~IEEE,}
\IEEEcompsocitemizethanks{\IEEEcompsocthanksitem P.-F. Marteau is with UMR CNRS IRISA, Université Bretagne Sud, F-56000 Vannes, France.}\protect\\
E-mail: see http://people.irisa.fr/Pierre-Francois.Marteau/
\thanks{}}

\IEEEtitleabstractindextext{%
\begin{abstract}
In the light of regularized dynamic time warping kernels, this paper re-considers the concept of time elastic centroid for a set of time series. We derive a new algorithm based on a probabilistic interpretation of kernel alignment matrices. This algorithm expresses the averaging process in terms of a stochastic alignment \textit{automata}. It uses an  iterative agglomerative heuristic method for averaging the aligned samples, while also averaging the times of occurrence of the aligned samples. By comparing classification accuracies for 45 heterogeneous time series datasets obtained by first nearest centroid/medoid classifiers we show that: i) centroid-based approaches significantly outperform medoid-based approaches, ii) for the considered datasets, our algorithm that combines averaging in the sample space and along the time axes, emerges as the most significantly robust model for time-elastic averaging with a promising noise reduction capability. We also demonstrate its benefit in an isolated gesture recognition experiment and its ability to significantly reduce the size of training instance sets. Finally we highlight its denoising capability using demonstrative synthetic data: we show that it is possible to retrieve, from few noisy instances, a signal whose components are scattered in a wide spectral band. 
\end{abstract}

}

\maketitle

\section{Introduction}
\label{intro}
Since Maurice Fr\'{e}chet's pioneering work \cite{frechet1906} in the early 1900s, \textit{time-elastic} matching of time series or symbolic sequences has attracted much attention from the scientific community in numerous fields such as information indexing and retrieval, pattern analysis, extraction and recognition, data mining, etc. This approach has impacted a very wide spectrum of applications addressing socio-economic issues such as the environment, industry, health, energy, defense and so on.

Among other time elastic measures, Dynamic Time Warping (DTW) was widely popularized during the 1970s with the advent of speech recognition systems
\cite{VelichkoZagoruyko1970}, \cite{SakoeChiba1971}, and numerous variants that have since been proposed to match time series with a certain degree of time distortion tolerance.

The main issue addressed in this paper is time series or shape averaging in the context of a time elastic distance. Time series averaging or signal averaging is a long-standing issue that is currently becoming increasingly prevalent in the big data context; it is relevant for de-noising \cite{Kaiser1979}, \cite{Hassan2010}, summarizing subsets of time series \cite{Petitjean2011}, defining significant prototypes, identifying outliers \cite{Gupta2014}, performing data mining tasks (mainly exploratory data analysis such as clustering) and speeding up classification \cite{Petitjean2014}, as well as regression or data analysis processes in a big data context.

In this paper, we specifically tackle the question of averaging subsets of time series, not from considering the DTW measure itself as has already been largely exploited, but from the perspective of the so-called regularized DTW kernel (KDTW). From this new perspective, the estimation of a time series average or centroid can be readily addressed with a probabilistic interpretation of kernel alignment matrices allowing a precise definition of the average of a pair of time series from the expected value of local alignments of samples. The tests carried out so far demonstrate the robustness and the efficiency of this approach compared to the state of the art approach.

The structure of this paper is as follows: the introductory section, the second section summarizes the most relevant related studies on time series averaging as well as DTW kernelization. In the third section, we derive a probabilistic interpretation of kernel alignment matrices evaluated on a pair of time series by establishing a parallel with a forward-backward procedure on a stochastic alignment automata. In the fourth section, we define the average of a pair of time series based on the alignment expectation of pairs of samples, and we propose an algorithm designed for the averaging of any subset of time series using a pairwise aggregating procedure. We present in the fifth section three complementary experiments to assess our approach against the state of the art, and conclude.

\section{Related works}
\label{sec:RelatedWorks}
Time series averaging in the context of (multiple) time elastic distance alignments has been mainly addressed in the scope of the Dynamic Time Warping (DTW) measure \cite{VelichkoZagoruyko1970}, \cite{SakoeChiba1971}. Although other time elastic distance measures such as the Edit Distance With Real Penalty (ERP) \cite{Chen04ERP} or the Time Warp Edit Distance (TWED) \cite{Marteau09TWED} could be considered instead, without loss of generality, we remain focused throughout this paper on DTW and its kernelization. 

\subsection{DTW and time elastic average of a pair of time series}
A classical formulation of DTW can be given as follows. If $d$ is a fixed positive integer, we define a time series of length $n$ as a multidimensional sequence $X_1^n=X_1X_2\cdots X_n$, such that, $\forall i \in \{1, ..,n\}$,  $X_i \in \mathbb{R}^d$.\\

\begin{definition}
\label{def:alignmentPath}
    If $X_1^n$ and $Y_1^m$ are two time series with respective lengths $n$ and $m$, an {\it alignment path} $\pi = (\pi_k)$ of length $p=|\pi|$ between $X_1^n$ and $Y_1^m$ is represented by a sequence
    \[
        \pi : \{1, \ldots, p\} \rightarrow \{1, \ldots, n\} \times \{1, \ldots, m\}
    \]
    such that $\pi_1 = (1, 1)$, $\pi_p = (n, m)$, and (using the
    notation $\pi_k = (i_k, j_k)$, for all $k \in \{1, \ldots, p-1\}$,
    $\pi_{k+1} = (i_{k+1}, j_{k+1}) \in \{(i_k + 1, j_k),\linebreak[1](i_k, j_k + 1),\linebreak[1](i_k + 1, j_k + 1) \}$.\\

We define $\forall k$ $\pi_{k}(1)=i_k$ and $\pi_{k}(2)=j_k$, as the index access functions at step $k$ of the mapped elements in the pair of aligned time series.\\
\end{definition}

    In other words, a warping path defines a way to travel along both time series simultaneously from beginning to end; it cannot skip a point, but it can advance one time step along one series without advancing along the other, thereby justifying the term \textit{time-warping}.

    If $\delta$ is a distance on $\mathbb{R}^d$, the global {\it cost}
    of a warping path $\pi$ is the sum of distances (or squared distances or local costs) between pairwise elements of the two time series along $\pi$, i.e.:
    \[
        \text{cost}(\pi) = \sum_{(i_k,j_k) \in \pi} \delta(X_{i_k}, Y_{j_k})
    \]
    A common choice of distance on $\mathbb{R}^d$ is the one generated by the
    $L^2$ norm.

\begin{definition}
\label{dtw}
    For a pair of finite time series $X$ and $Y$, any warping path has a finite length, and thus the number of existing warping paths is finite. Hence, there exists at least one path $\pi^*$ whose cost is minimal, so we can define $\text{DTW}(X, Y)$ as the minimal cost taken over all existing warping paths. Hence
\begin{align}
\label{eq:dtw}
      \text{DTW}(X_1^n, Y_1^m) &= \underset{\pi}{\min} \text{ cost}(\pi(X_1^n, Y_1^m))\nonumber \\
      &=\text{cost}(\pi^*(X_1^n, Y_1^m)).
\end{align}

\end{definition}

\begin{definition}
\label{pairwiseTSaverage}
From the DTW measure, \cite{Gupta1996}~ have defined the time elastic average  $a(X,Y)$ of a pair of time series $X_1^n$ and $Y_1^m$ as the time series $A_1^{|\pi^*|}$ whose elements are $A_k=\textsl{mean}(X_{\pi^*_k(1)}, Y_{\pi^*_k(2)})$, $\forall k \in {1, \cdots, |\pi^*|}$, where \textsl{mean} corresponds to the definition of the mean in Euclidean space.\\
\end{definition}

  \begin{figure*}[!ht]
    \subfloat[Pairwise average (top) and Progressive agglomeration (bottom)\label{fig:hac-iter-a}]{%
    \fbox{\includegraphics[scale=.37]{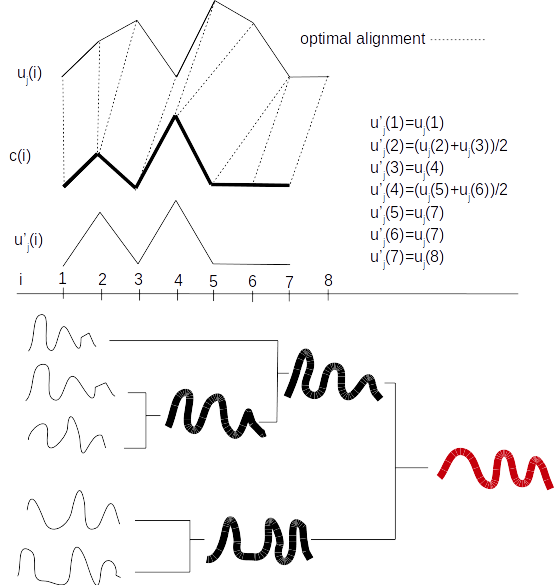}}\\
    }
    \hfill
    \subfloat[Iterative agglomeration with refinement\label{fig:hac-iter-b}]{%
      	\fbox{\includegraphics[scale=.37]{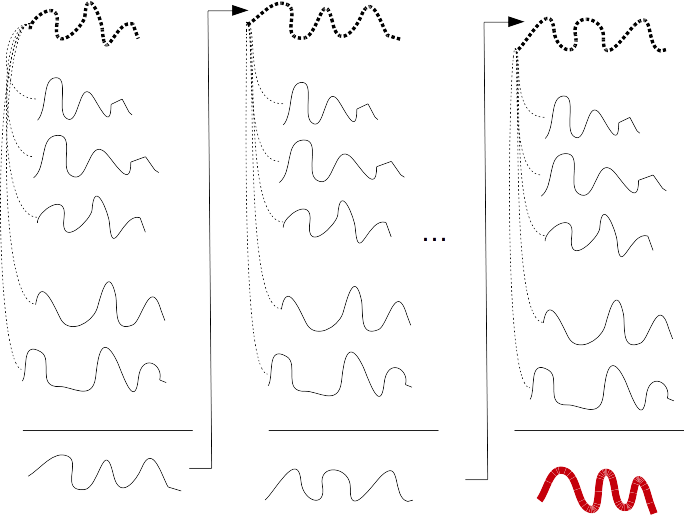}}
    }
    \caption{Pairwise averaging (top left), progressive hierarchical with similar first agglomeration (bottom left) v.s. iterative agglomeration (right) strategies. Final centroid approximations are presented in red bold color. Temporary estimations are presented using a bold dotted black line}
    \label{fig:hac-iter}
  \end{figure*}

\subsection{Time elastic centroid of a set of time series}
A single alignment path is required to calculate the time elastic centroid of a pair of time series (Def. \ref{def:alignmentPath}). However, multiple path alignments need to be considered to evaluate the centroid of a larger set of time series. Multiple alignments have been widely studied in bioinformatics \cite{Fasman1998}, and it has been shown that determining the optimal alignment of a set of sequences under the sum of all pairs (SP) score scheme is a NP-complete problem \cite{WangJ1994} \cite{Just99}. The time and space complexity of this problem is $O(L^k)$, where $k$ is the number of sequences in the set and $L$ is the length of the sequences when using dynamic programming to search for an optimal solution \cite{Carrillo1988}. This latter result applies to the estimation of the time elastic centroid of a set of $k$ time series with respect to the DTW measure. Since the search for an optimal solution becomes rapidly intractable with increasing $k$, sub-optimal heuristic solutions have been subsequently proposed, most of them falling into one of the following three categories.

\subsubsection{Progressive heuristics}
Progressive heuristic methods estimate the time elastic centroid of a set of $k$ time series by combining pairwise centroids (Def. \ref{pairwiseTSaverage}). This kind of approach constructs a binary tree whose leaves correspond to the time series of the data set, and whose nodes correspond to the calculation of a local pairwise centroid, such that, when the tree is complete, the root is associated with the estimated data set centroid. The proposed strategies differ in the way the tree is constructed. One popular approach consists of providing a random order for the leaves, and then constructing the binary tree up to the root using this ordering \cite{Gupta1996}. Another approach involves constructing a dendrogram (a hierarchical ascendant clustering) from the data set and then using this dendrogram to calculate pairwise centroids starting with the closest pairs of time series and progressively aggregating series that are farther away\cite{Niennattrakul2009} as illustrated on the left of Figure \ref{fig:hac-iter}. Note that these heuristic methods are entirely based on the calculation of a pairwise centroid, so they do not explicitly require the evaluation of a DTW centroid for more than two time series. Their degree of complexity varies linearly with the number of time series in the data set. 

\subsubsection{Iterative heuristics}
Iterative heuristics are based on an iterated three-step process. For a given temporary centroid candidate, the first step consists of calculating the inertia, i.e. the sum of the DTW distances between the temporary centroid and each time series in the data set. The second step (Figure \ref{fig:hac-iter-a} top) evaluates the best pairwise alignment with the temporary centroid $c(i)$, of length $L$, for each time series $u_j(i)$ in the data set ($j \in \{1 \cdots n\}$), where $i$ is the timestamp. A new time series of length $L$, $u'_j(i)$ is thus constructed that contains the contributions of all the samples of time series $u_j(i)$, but with  time being possibly stretched (duplicate samples) or compressed (average of successive samples) according to the best alignment path as exemplified in Figure \ref{fig:hac-iter-a}, top left side. The third step consists in producing a new temporary centroid candidate $c(i)$ from the set $\{u'_j(i)\}$ by successively averaging (in the sense of the Euclidean centroid), the samples at every timestamp $i$ of the $u'_j(i)$ time series. Basically, we have $c(i) = 1/n\cdot\sum_{j=1..n} u'_j(i)$. 

Then, the new centroid candidate replaces the previous one and the process is iterated until the inertia is no longer reduced or the maximum number of iterations is reached. Generally, the first temporary centroid candidate is taken as the DTW medoid of the considered data set. This process is illustrated on Figure \ref{fig:hac-iter-b}. The three steps of this heuristic method were first proposed in \cite{Abdulla2003}. The iterative aspect of this heuristic approach was initially introduced by \cite{Hautamaki2008} and refined by \cite{Petitjean2011} who introduced the DTW Barycenter Averaging (DBA) algorithm. Note that, in contrast to the progressive method, this kind of approach needs to evaluate, at each iteration, all the alignments with the current centroid candidate. The complexity of the iterative approach is higher than the progressive approach, the extra computational cost being linear with the number of iterations. More sophisticated approaches have been proposed to escape some local minima. For instance \cite{Petitjean2012} have evaluated a genetic algorithm for managing a population of centroid candidates,  thus improving with some success the straightforward iterative heuristic methods.    

\subsubsection{Optimization approaches} 
Given the entire set of time series $\mathbb{S}$ and a subset of $n$ time series $S=\{X_j\}_{j=1 \cdots n} \subseteq \mathbb{S}$, optimization approaches attempt to estimate the centroid of $S$ from the definition of an optimization problem, which is generally expressed by equation (\ref{eq.opt}) given below: 
\begin{equation}
\label{eq.opt}
c = \argmin_{s \in \mathcal{S}} \sum_{j=1}^n \DTW(s,X_j) 
\end{equation}

%\color{red}
Among other works, some attempt to use this kind of direct approach for the estimation of time elastic centroid was recently addressed in \cite{ZhouDLTore2009}, \cite{ZhouDLTore2016} and \cite{SoheilyKhah2016}.

In \cite{ZhouDLTore2009} the authors detail a Canonical Time Warp (CTW) and a Generalized version of it (GCTW) \cite{ZhouDLTore2016} that combines DTW and CCA (Canonical Correlation Analysis) for temporally aligning multi-modal motion sequences. From a least square formulation for DTW, a non-convex optimization problem is handled by means of a coordinate-descent approach that alternates between multiple temporal alignments using DTW (or a variant exploiting a set of basis functions to parameterized the warping paths) and spatial projections using CCA (or a multi-set extension of CCA). Whilst these approaches have not been designed to explicitly propose a centroid estimation, they do provide multi-alignment paths that can  straightforwardly be used to compute a centroid estimate. As an extension to CTW, GCTW requires the set-up of generally "smooth" function basis that constrain the shape of the admissible alignment paths. This ensures the computational efficiency of GCTW, but in return it may induce some drawback,  especially when considering the averaging of "unsmoothed" time series that may involve very "jerky" alignment paths. The choice of this function basis may require some expertise on the data. \\

\color{black}
In \cite{SoheilyKhah2016},  a non-convex constrained optimization problem is derived, by integrating a temporal weighting of local sample alignments to highlight the temporal region of interest in a time series data set, thus penalizing the other temporal regions. 
%Two time elastic measures were specifically addressed: i) a dynamic time warping measure between a time series and a weighted time series (representing the centroid estimate) and ii) a DTW based kernel called DTAK \cite{Shimodaira2002}. Their results are very promising: 
Although the number of parameters to optimize is linear with the size and the dimensionality of the time series, the two steps gradient-based optimization process they derived is very computationally efficient and shown to outperform the state of the art approaches on some challenging scalar and multivariate data sets. However, as numerous local \textit{optima} exist in practice, the method is not guaranteed to converge towards the best possible centroid, which is anyway the case in all other approaches. Furthermore, their approach, due to  combinatorial explosion, cannot be adapted for time elastic kernels like the one addressed in this paper and described in section \ref{sec:DTW}.

\subsection{Discussion and motivation}
According to the state of the art in time elastic centroid estimation, an exact centroid, if it exists, can be calculated by solving a NP-complete problem whose complexity is exponential with the number of time series to be averaged. Heuristic methods with increasing time complexity have been proposed since the early 2000s. Simple pairwise progressive aggregation is a less complex approach, but which suffers from its dependence on initial conditions. Iterative aggregation is reputed to be more efficient, but entails a higher computational cost. It could be combined with ensemble methods or soft optimization such as genetic algorithms. The non-convex optimization approach has the merit of directly addressing the mathematical formulation of the centroid problem in a time elastic distance context. This approach nevertheless involves a higher complexity and must deal with a relatively large set of parameters to be optimized (the weights and the sample of the centroid). Its scalability could be questioned, specifically for high dimensional multivariate time series.   

It should also be mentioned that some criticism of these heuristic methods has been made in \cite{Niennattrakul2007}. Among other drawbacks, the fact that DTW is not a metric %(the triangle inequality is not satisfied) 
could explain the occurrence of unwanted behavior such as centroid drift outside the time series cluster to be averaged. We should also bear in mind that keeping a single best alignment %(even though several may exist, without mentioning the \textit{good} ones) 
can increase the dependence of the solution on the initial conditions. It may also increase the aggregating order of the time series proposed by the chosen method, or potentially enhance the convergence rate.    

In this study, we do not directly address the issue of time elastic centroid estimation from the DTW perspective, but rather from the point of view of the regularized dynamic time warping kernel (KDTW) \cite{MarteauGibet2014}. Although this perspective allows us to consider centroid estimation as a preimage problem, which is in itself another optimization perspective, we rather show that the KDTW alignment matrices computation can be described as the result of applying a forward-backward algorithm on a stochastic alignment automata.  This probabilistic interpretation of the pairwise alignment of time series leads us to propose a robust averaging scheme for any set of time series that interpolate jointly along the time axis and in the sample space.  Furthermore, this scheme significantly outperforms the current state of the art method, as shown by our experiments. 

\subsection{Time elastic kernels and their regularization}    
\label{sec:DTW}
The \textbf{Dynamic Time Warping} (DTW) distance between two time series $X_1^p=X_1X_2 \cdots X_p$ and $Y_1^q = Y_1Y_2 \cdots Y_q $ of lengths $p$ and $q$ respectively, \cite{VelichkoZagoruyko1970}, \cite{SakoeChiba1971} as defined in equation (\ref{eq:dtw}) can be recursively evaluated as 

\begin{eqnarray}
\label{Eq.dtw2}
 d_{dtw}(X_1^p, Y_1^q)= \hspace{5cm} \nonumber \\
  d_{E}^{2}(X_p, Y_q)  
  + \text{Min} 
   \left\{
   \begin{array}{ll}
     d_{dtw}(X_1^{p-1}, Y_1^q)  \\
     d_{dtw}(X_1^{p-1}, Y_1^{q-1})  \\ 
     d_{dtw}(X_1^p, Y_1^{q-1})   \\
   \end{array} \right.
\end{eqnarray}
where $d_{E}(X_p,Y_q)$ is the Euclidean distance defined on $\mathbb{R}^d$ between the two positions in sequences $X_1^p$ and $Y_1^q$ taken at times $p$ and $q$, respectively. 
 
Apart from the fact that the triangular inequality does not hold for the DTW distance measure, it is not possible to define a positive definite kernel directly from this distance. Hence, the optimization problem, which is inherent to the learning of a kernel machine, is no longer convex and could be a source of limitation due to the emergence of local minima.\\

\textbf{Regularized DTW}: seminal work by  \cite{CuturiVert2007}, prolonged recently by \cite{MarteauGibet2014} leads us to propose new guidelines to ensure that kernels constructed from elastic measures such as DTW are positive definite. A simple instance of such a regularized kernel, derived from \cite{MarteauGibet2014}, can be expressed as a convolution kernel, which makes use of two recursive terms:

%\color{red}
%\begin{equation}
%
\resizebox{.95\linewidth}{!}{
  \begin{minipage}{\linewidth}
  \begin{align}
  \label{Eq.KDTW}
\begin{array}{ll}
\textsc{KDTW} (X_1^p, Y_1^q)=K_{dtw}(X_1^p, Y_1^q)+K'_{dtw}(X_1^p, Y_1^q) \\
\\
K_{dtw}(X_1^p, Y_1^q) =  \\
   \begin{array}{ll}
    \hspace{2mm} \frac{1}{3}e^{-\nu d_{E}^{2}(X_p, Y_q)} \cdot \Big(h(p-1,q)K_{dtw}(X_1^{p-1}, Y_1^q)\\
   \hspace{20mm}  + h(p-1,q-1) K_{dtw}(X_1^{p-1}, Y_1^{q-1})  \\
   \hspace{20mm}  + h(p,q-1)K_{dtw}(X_1^p, Y_1^{q-1})\Big) \\
   \end{array}\\
   {}\\
   K'_{dtw}(X_1^p, Y_1^q) =  \\
   \begin{array}{ll}
    \hspace{2mm} \frac{1}{3} \cdot \Big(h(p-1,q) K'_{dtw}(X_1^{p-1}, Y_1^q)e^{-\nu d_{E}^{2}(X_p, Y_p)}  \\
    \hspace{3mm} + \Delta_{p,q} h(p-1,q-1)K'_{dtw}(X_1^{p-1}, Y_1^{q-1})e^{-\nu d_{E}^{2}(X_p, Y_q)}\\
    \hspace{3mm}   + h(p,q-1)K'_{dtw}(X_1^p, Y_1^{q-1})e^{-\nu d_{E}^{2}(X_q, Y_q)}\Big) \\
   \end{array}
  \end{array}
\end{align}
 \end{minipage}
}
\color{black}

where $\Delta_{p,q}$ is the Kronecker symbol, $\nu \in \mathbb{R}^{+}$ is a \textit{stiffness} parameter which weights the local contributions, i.e. the distances between locally aligned positions, $d_E(.,.)$ is a distance defined on $\mathbb{R}^{k}$, and $h$ is a symmetric binary non negative function, usually in $\{0,1\}$, used to define a symmetric corridor around the main diagonal to limit the "time elasticity" of the kernel. For the remaining of the paper we will not consider any corridor, hence $h(.,.)=1$ everywhere.

The initialization is simply $K_{dtw}(X_1^0, Y_1^0) = K'_{dtw} (X_1^0, Y_1^0) = 1$.\\

The main idea behind this regularization is to replace the operators $\min$ and $\max$ (which prevent symmetrization of the kernel) by a summation operator. This allows us to consider the best possible alignment, as well as all the best (or nearly the best) paths by summing their overall cost. The parameter $\nu$ is used to check what is termed as nearly-the-best alignment, thus penalizing alignments that are too far away from the optimal ones. This parameter can be easily optimized through a cross-validation. \\

%\color{red}
For each alignment path, KDTW evaluates the product of local alignment costs  $e^{-\nu d_{E}^{2}(X_p, Y_q))} \le 1 $ occurring along the path. This product  can be very small depending on the size of the time series and the selected value for $\nu$. This is the source for a diagonal dominance problem in the Gram matrix.  But, above all, this requires to balance the choice of the $\nu$ value according to the lengths of the matched time series. This is the main (and probably the only) limitation of the KDTW kernel: the selectivity or bandwidth of the local alignment kernels needs to be adjusted according to the lengths of the matched time series. 

\section{Stochastic alignment process}

To introduce a probabilistic paradigm to the time elastic averaging of time series, we first consider the pairwise alignment process as the output of a stochastic automata. The stochastic alignment process that we propose finds its roots in the forward-backward algorithm defined for the learning of Hidden Markov Models (HMM) \cite{Rabiner89} and in the parallel between HMM and DTW that is proposed in \cite{Juang85}, \cite{NakagawaNakanishi1989}  and in a more distant way in \cite{Chudova2002}. However we differ from these founding works (and others) in the following
\begin{enumerate}
\item we do not construct a parallel with DTW, but with its kernelized variant KDTW
\item \cite{NakagawaNakanishi1989} only consider an optimal alignment path (exploiting the Viterbi algorithm) while we consider the whole set of possible alignments (as in \cite{Juang85})
\item \cite{Juang85} construct an asymmetric classical left-right HMM (one time series plays the role of the observation sequence, while the other plays the role of the state sequence). With a similar idea \cite{Chudova2002} proposes a generative mixture model along a discrete time grid axis with local and global time warp capability. We construct instead an alignment process, that sticks on the DTW recursive definition without any other hypothesis on the structure of the automata, and for which the two aligned time series play the role of the observation sequence, and the set of states corresponds to the set of all possible sample pairs alignments. 
\end{enumerate}

\subsection{pairwise alignment of time series as a Markov model}
Let $o_1^n = o_1 o_2 \cdots o_n$ and ${o'}_1^{n'} = o'_1 o'_2 \cdots  o'_{n'}$ be two discrete time series (observations) of length $n$ and $n'$ respectively. To align this two time series, we define a stochastic alignment automata as follows. First we consider the set of state variables $\mathcal{S} =\{ S_{1,1},S_{1,2}, \cdots, S_{n,n'}\}$. Each $S_{i,j}$ characterizes the alignment between observed samples $o_i$ and $o'_j$. The posterior probability for all state variables, $S_{i,j}$, given the sequences of observations $o_1^n$ and ${o'}_1^{n'}$ is $P(S_{i,j}|o_1^n;{o'}_1^{n'})$.

The transitions probabilities between states are driven by a tensor $\mathbf{A}=[a_{ij;kl}]$, where $a_{ij;kl}=P(S_{k,l}|S_{i,j})$, $\forall (k,l)$ and $(i,j) \in \{1 \cdots n\} \times \{1 \cdots n'\}$. $\mathbf{A}$ can be defined accordingly to the standard DTW definition, namely
\begin{equation}
\label{Atensor}
a_{ij;kl} = \left\{
\begin{array}{ll}
\frac{1}{3} \textsc{ if } \left\{
\begin{array}{ll}
(k=i \textsc{ and } l=j+1) \\
\textsc{ or } (k=i+1 \textsc{ and } l=j+1) \\
\textsc{ or } (k=i+1 \textsc{ and } l=j)\\
\end{array} 
\right. \\
0 \textsc{ otherwise.}\\
\end{array}
\right.
\end{equation}
The $1/3$ factor ensures that the transition matrix equivalent to $\mathbf{A}$ is stochastic, basically 
\begin{equation}
\label{eq:stochastic}
\forall i,j \textsc{ }\sum_{kl} a_{ij;kl} = 1\\
\end{equation}
Notice that any tensor $\mathbf{A}$ satisfying equation (\ref{eq:stochastic}) could be considered at this level instead of the previous DTW surrogate tensor.\\

Furthermore, each state is observable through the so-called emission probabilities which are defined by a set of functions $b_{ij}(x,y)$, where $b_{ij}(x,y)=P(x,y | S_{i,j})$, $\forall (x,v) \in \mathbb{R}^d \times \mathbb{R}^d$ and $(i,j) \in \{1 \cdots n\} \times \{1 \cdots n'\}$. The $b_{ij}$ functions are normalized such that $\iint_{x,y} b_{ij}(x,y) \,dx\,dy=1$. \\
%In our automata framework, these emission probabilities are independent of the state the 

%Here we differ from the classical HMM: for our construction, the states are not truly hidden since the knowledge of the local observation pair (a local alignment $(o_u,o'_v)$) fully determines the state ($S_{uv}$).
Here we differ from the classical HMM: 
 the first difference lies in the nature of the observation sequence itself. Unlike HMM, our observation consists of a pair of subsequences that are not traveled necessarily synchronously, but according to the structure of the transition tensor $\mathbf{A}$. For instance, given the DTW tensor described by equation (\ref{Atensor}), from a current state associated to the alignment $(o_u,o'_v)$, three possible alignments can be reached at the next transition: $(o_{u+1},o'_v)$, $(o_u,o'_{v+1})$ or $(o_{u+1},o'_{v+1})$.\\
 
The second difference with classical HMM is that the emission probabilities are independent from the state, such that $\forall i,j$ $b_{i,j}(x,y)=b(x,y)$. We use a local (density) kernel to estimate these probabilities as follows

\begin{equation}
\label{Btensor}
b(x,y) = \kappa(x,y)  = \gamma e^{-\nu d_{E}^{2}(x,y)}
\end{equation} 
where $\gamma$ is the normalization coefficient.

Consequently, given the two observation sequences  ${o}_1^{n}$ and ${o'}_1^{n'}$, we define the emission probability matrix $\mathbf{B}=[b_{kl}]=b(o_k,o'_l)=\gamma e^{-\nu d_{E}^{2}(o_k,o'_l)}$, for $k \in \{1,\cdots,n\}$ and $l \in \{1,\cdots,n'\}$\\

%where $0\leq \kappa(o_u, o'_v)\leq 1$ is any density kernel or discrete distribution measure (e.g. $\kappa(.,.) \propto e^{-\nu d_{E}^{2}(.,.)}$ for KDTW). \\

Finally let $\mathbf{u}$ be the initial state probability vector defined by $\forall (i,j) \in \{1 \cdots n\} \times \{1 \cdots n'\}$, $\mathbf{u}_{ij}=1$ if $i=j=1$, $0$ otherwise. \\

Thereby, the stochastic alignment automata is fully specified by the triplet $\theta=(\mathbf{A}, \mathbf{B}, \mathbf{u})$, where $\mathbf{A}$ only depends on the lengths $n$ and $n'$ of the observations, and $\mathbf{B}$ depends on the complete pair of observations $o_1^n$ and ${o'}_1^{n'}$.

\subsection{Forward-backward alignment algorithm}
We derive the forward-backward alignment algorithm for our stochastic alignment automata from its classical derivation that was defined for Hidden Markov Models \cite{Rabiner89}.

For all $S \in \mathcal{S}$, the posterior probability  $P(S|o_1^n,{o'}_1^{n'}, \theta)$ is decomposed into forward/backward recursions as follows:

\begin{equation}
\label{FB1}
\begin{array}{ll}
P(S|o_1^n,{o'}_1^{n'},\theta) &= \frac{P(o_1^n,{o'}_1^{n'}, S |\theta)}{P(o_1^n,{o'}_1^{n'}|\theta)} \\
&= \frac{P(o_1^t,o_t^n,{o'}_1^{t'},{o'}_{t'}^{n'},S|\theta)}{P(o_1^n,{o'}_1^{n'}|\theta)}\\
&= \frac{P(o_t^n,{o'}_{t'}^{n'}|S,\theta)P(S,o_1^t,{o'}_{1}^{t'}|\theta)}{P(o_1^n,{o'}_1^{n'}|\theta)}\\
\end{array}
\end{equation} 

The last equality results from the application of the Bayes rule and the conditional independence of $o_t^n,{o'}_{t'}^{n'}$ and $o_1^t,{o'}_{1}^{t'}$ given $S$, $\theta$.\\

Let $\alpha_{t,t'}=P(o_1^t,{o'}_{1}^{t'},S_{t,t'}|\theta)$ be the probability of the alignment of the pair of partial observation sequences $(o_1^t, {o'}_{1}^{t'})$ produced by all possible state sequences that end at state $S_{t,t'}$.  $\alpha_{t,t'}$ can be recursively evaluated as the forward procedure
\begin{equation}
\label{Forward}
\left\{
\begin{array}{ll}
\alpha_{1,1}=u_{11}b_{11}\\
\alpha_{t,t'}=b_{tt'}\sum\limits_{u,v\in \mathcal{F}_{t,t'}}\alpha_{u,v}a_{uv;tt'} \\
\end{array}
\right.
\end{equation} 
where $\mathcal{F}_{t,t'}$ is the subset of states allowing to reach the state $S_{t,t'}$ in a single transition. For the DTW tensor $\mathbf{A}$ (Eq.  \ref{Atensor}), we have $\mathcal{F}_{t,t'}=\{S_{t-1,t'}, S_{t,t'-1}, S_{t-1,t'-1}\}$.\\
Notice that in this case $\alpha_{n,n'}=K_{dtw}(o_1^n, {o'}_{1}^{n'})$. \\

Similarly let $\beta_{t,t'}=P(o_t^n,{o'}_{t'}^{n'}|S,\theta)$ be the probability of the alignment of the pair of partial sequences $(o_t^n, {o'}_{t'}^{n'})$ given starting state $S_{t,t'}$. $\beta_{t,t'}$ can be recursively evaluated as the backward procedure
\begin{equation}
\label{Backward}
\left\{
\begin{array}{ll}
\beta_{n,n'}=1\\
\beta_{t,t'}=\sum\limits_{u,v\in \mathcal{B}_{t,t'}}\beta_{u,v}a_{tt';uv}b_{tt'} \\
\end{array}
\right.
\end{equation}

where $\mathcal{B}_{t,t'}$ is the subset of states that can be reached from the state $S_{t,t'}$ in a single transition. For the DTW tensor $\mathbf{A}$ (Eq. \ref{Atensor}), we have $\mathcal{B}_{t,t'}=\{S_{t+1,t'}, S_{t,t'+1}, S_{t+1,t'+1}\}$.\\

Hence from Eq. \ref{FB1}, we get 
\begin{equation}
\label{FB2}
P(S_{t,t'}|o_1^n,{o'}_1^{n'},\theta) = \frac{\alpha_{t,t'}\beta_{t,t'}}{P(o_1^n,{o'}_1^{n'}|\theta)}
\end{equation} 

Any tensor $\mathbf{A}$ satisfying equation (\ref{eq:stochastic}) is not eligible: for the $\alpha_{t,t'}$ and $\beta_{t,t'}$ recursions to be calculable, one has to impose \textit{linearity}. Basically $\alpha_{t,t'}$ cannot depend on any $\alpha_{u,v'}$ that is not previously evaluated. The constraint we need to impose is that the time stamps are locally increasing, i.e. if  $\alpha_{t,t'}$ depends on any $\alpha_{u,v'}$, then necessarily $[(t<u $ and $t'\leq v')$ or $(t\leq u $ and $ t' < v')]$. The same applies for the $\beta_{t,t'}$ recursion.\\

\begin{figure}[h!]
\centering
\begin{tabular}{ccc}
	\includegraphics[scale=.6, angle=0]{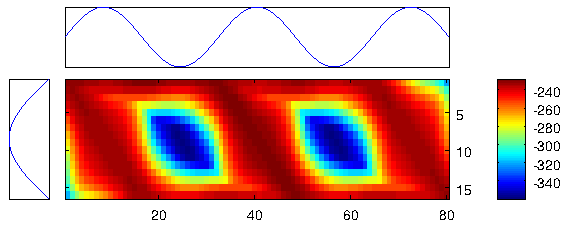} &
\end{tabular}
\caption{Forward Backward matrix (logarithmic values) for the alignment of a positive halfwave with a sinus wave. The dark red color represents high probability states, while dark blue color represents low probability states.}
\label{fig:SinTest}
\end{figure}

As an example, Figure \ref{fig:SinTest} presents the Forward Backward ($FB$) matrix ($FB(t,t')=P(S_{t,t'}|o_1^n,{o'}_1^{n'},\theta)$)  corresponding to the alignment of a positive half-wave with a sinus wave. The three areas of likely alignment paths are clearly identified in dark red colors. \\

\subsection{Parallel with KDTW}
A direct parallel exists between KDTW and the previous Markov process. It follows from the forward equation (Eq. \ref{Forward}) that 

\begin{align}
\label{KDTW_HMM}
K_{dtw}(X_1^k, Y_1^l)=\sum_{i,j} a_{ij,kl} b_{kl}K_{dtw}(X_1^i), Y_1^j) \nonumber\\
  =\kappa(X_k,Y_l) \sum_{i,j} a_{ij,kl} K_{dtw}(X_1^i, Y_1^j)
\end{align} 

where $\mathbf{A}=[a_{ij;kl}]$ is defined in equation (\ref{Atensor}), and $\mathbf{B}=[b_{kl}]$, defined in equation (\ref{Btensor}), is such that $b_{kl}=e^{-\nu d_{E}^{2}(X_k,Y_l)}$. Hence, the $K_{dtw}$ recursion coincides exactly with the forward recursion (Eq. \ref{Forward}).
Similarly, we can assimilate the backward recursion (eq. \ref{Backward}) to the $K_{dtw}$ evaluation of the pair of time series obtained by inverting $X$ and $Y$ along the time axis. Hence, the forward-backward matrix elements (eq. \ref{FB2}) can be directly expressed in terms  $K_{dtw}$ recursions.

Furthermore, the corridor function $h()$ that occurs in the $K_{dtw}$ recursion (Eq. \ref{Eq.KDTW}) modifies directly the structure of the transition tensor  
 $\mathbf{A}$ by setting $a_{ij;kl}=0$ whenever $h(i,j)=0$ or $h(k,l)=0$. Neighbor states may be affected also by the normalization that is required to maintain  $\mathbf{A}$ stochastic.
 
%\subsection{Time elastic averaging of a pair of time series}
\subsection{Time elastic centroid estimate of a set of time series}

Let us introduce the marginal probability of subset $S_{t,\bullet}=\{S_{t,1}, S_{t,2}, \cdots, S_{t,n'}\}$ given the observations $o$ and $o'$, namely that sample $o_t$ is aligned with the samples of ${o'}_{1}^{n'}$
\begin{equation}
P(S_{t,\bullet})=\sum_{t'}P(S_{t,t'}|o_1^n,{o'}_{1}^{n'},\theta)
\end{equation}

and let us consider, for all $t$ and $t'$, the conditional probability of state $S_{t,t'}$ given the two observation sequences, parameter $\theta$ and $S_{t,\bullet}$, namely the probability that $o_t$ and  $o'_{t'}$ are aligned given the knowledge that $o_t$ is aligned with one of the samples of $o'$.
\begin{equation}
\label{CondProp1}
\begin{array}{ll}
P(S_{t,t'}|o_1^n,{o'}_{1}^{n'}, S_{t,\bullet}, \theta) = \\
 \hspace{5mm}P(S_{t,t'}|o_1^n,{o'}_{1}^{n'},\theta)/P(S_{t,\bullet}|o_1^n,{o'}_1^{n'},\theta)
\end{array}
\end{equation} 

The previous equality is easily established because $P(S_{t,t'}, S_{t,\bullet}|o_1^n,{o'}_{1}^{n'}, \theta) = P(S_{t,t'}|o_1^n,{o'}_{1}^{n'}, \theta)$.\\

Note that for estimating $P(S_{t,t'}, S_{t,\bullet}|o_1^n,{o'}_{1}^{n'}, \theta)$ we only need to evaluate the forward ($\alpha_{t,t'}$) and backward ($\beta_{t,t'}$) recursions, since $P(o_1^n,{o'}_1^{n'}|\theta)$, the numerator term  in Eq.\ref{FB2}, is eliminated.\\

We can then define the expectation of the samples of ${o'}_{1}^{n'}$ that are aligned with sample  $o_t$ (given that $o_t$ is aligned) as well as the expectation of time of occurrence of the samples of  ${o'}_{1}^{n'}$ that are aligned with $o_t$ as follows:

\begin{equation}
\label{Expectation1} 
\begin{array}{ll}
E(o'|o_t)=\frac{1}{n'}\sum\limits_{t'=1}^{n'} {o'}_{t'} P(S_{t,t'}|o_1^n,{o'}_1^{n'},S_{t,\bullet},\theta)\\
E(t'|o_t)=\frac{1}{n'}\sum\limits_{t'=1}^{n'} {t'} P(S_{t,t'}|o_1^n,{o'}_1^{n'},S_{t,\bullet},\theta)\\
\end{array}
\end{equation}

\begin{figure}[h]
\centering
\begin{tabular}{ccc}
	\includegraphics[scale=.45, angle=0]{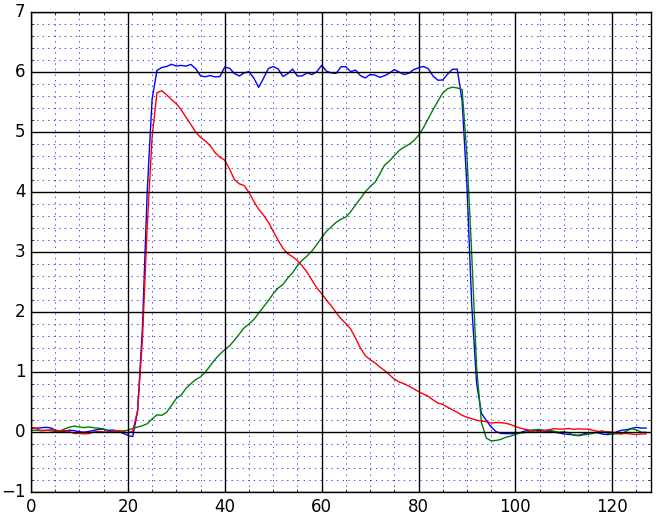} &
\end{tabular}
\caption{Centroids obtained for the CBF data set. For the three shapes, the expected start (24) and end (88) time stamps (hence the expected shape duration of 64 frames) are correctly extracted}
\label{fig:CBF_Cent}
\end{figure}
%
%To average a set of time series, the pairwise averaging principle described by equation (\ref{eq:PWA}) can be used to design a pairwise progressive agglomeration algorithm, as depicted in figure (\ref{fig:hac-iter}-a). We do not develop furthermore this algorithm and adopt instead an iterative approach that is described hereinafter.

%\subsection{Time elastic centroid estimate of a set of time series}

The Expectation equations  (Eq. \ref{Expectation1}) are at the basis of our procedure for averaging a set of time series. 

Let $O=\{{}^k\! o_1^{n_k}\}_{k=1 \cdots N}$ be a set of time series and $r_1^{n}$ a reference time series ($r_1^{n}$ can be initially setup  as the medoid of set $O$).
The centroid estimate of $O$ is defined as the pair  $(c, \tau)$ where $c$ is a time series of length $n$ and $\tau$ is the sequence of time stamps associated to the samples of $c$
%\begin{equation}
%\label{eq:TEKA}
%\begin{array}{ll}
%c(i)=\frac{1}{N} \sum\limits_{k=1}^N \sum\limits_{j=1}^{n_k} {}^k\! o_j \hat{P}(S_{ij}|r_1^{n},{}^k\! o_1^{n_k})\\
%\tau(i)=\frac{1}{N} \sum\limits_{k=1}^N \sum\limits_{j=1}^{n_k} j \hat{P}(S_{ij}|r_1^{n};{}^k\! o_1^{n_k})\\
%\end{array}
%\end{equation} 

\begin{equation}
\label{eq:TEKA}
\begin{array}{ll}
c(t)=\frac{1}{N} \sum\limits_{k=1}^N E({}^k\! o|r_t)\\
\hspace{7mm} =\frac{1}{N} \sum\limits_{k=1}^N \frac{1}{n_k} \sum\limits_{{}^k\! t=1}^{n_k} {}^k\! o_j P(S_{t,{}^k\! t}|r_1^{n},{}^k\! o_1^{n_k}, S_{t,\bullet})\\
\tau(t)=\frac{1}{N} \sum\limits_{k=1}^N E({}^k\! t|r_t)\\
\hspace{7mm} =\frac{1}{N} \sum\limits_{k=1}^N \frac{1}{n_k} \sum\limits_{{}^k\! t=1}^{n_k} {}^k\! t P(S_{t,{}^k\! t}|r_1^{n},{}^k\! o_1^{n_k}, S_{t,\bullet})\\
\end{array}
\end{equation} 

Obviously, $(c,\tau)$ is a non uniformly sampled time series for which $\tau(t)$ is the time stamp associated to observation $c(t)$. $\tau(t)$ could be understood as the  expected time of occurrence of the expected observation $c(t)$. A uniform re-sampling can straightforwardly be used to get back to a uniformly sampled time series.\\

The proposed iterative agglomerative algorithm (cf. Fig. \ref{fig:hac-iter}-b), called TEKA (Time Elastic Kernel Averaging), that provides a refinement of the centroid estimation at each iteration until reaching a (local) optimum is presented in algorithm (\ref{alg:TEKA}).\\

As an example, figure (\ref{fig:CBF_Cent}) presents the time elastic centroid  estimates obtained, using algorithm (\ref{alg:TEKA}) with $K=K_{dtw}$, for the Cylinder c(t), Bell, b(t) Funnel, f(t), synthetic functions \cite{Saito1994} defined as follows\\

$c(t) = (6 + \eta)\cdot\chi_{[a,b]}(t) + \epsilon(t)$\\
\hspace*{4mm} $b(t) = (6 + \eta)\cdot\chi_{[a,b]}(t)\cdot (t-a)/(b-a) + \epsilon(t)$\\
\hspace*{4mm} $f(t) = (6 + \eta)\cdot \chi_{[a,b]}(t) \cdot (b-t)/(b-a) + \epsilon(t)$\\
where $\chi_{[a,b]} =0$ if $t < a \vee t > b$, $1$ if $a \le t \le b$, 
$\eta$ and $\epsilon(t)$ are obtained from a standard normal distribution $N(0, 1)$, $a$ is an
integer obtained from a uniform distribution in $[16, 32]$ and $b-a$ is another
integer obtained from another uniform distribution in $[32, 96]$. 
Hence such shapes are characterized with start and end time stamps of $24$ and $88$ respectively, and a shape duration of $64$ samples.  Figure (\ref{fig:CBF_Cent}) clearly shows that, from a subset of 300 time series (100 for each category),  the algorithm has correctly recovered the start and end shape events (hence the expected shape duration) for all three shapes.

    \begin{algorithm}[]
      \caption{Iterative Time Elastic Kernel Averaging (TEKA) of a set of time series}\label{alg:TEKA}
      \begin{algorithmic}[1]
        \State Let $K$ be a similarity time elastic kernel for time series
         satisfying eq. (\ref{KDTW_HMM}) 
        \State Let $O$ be a set of time series of $d$ dimensional samples
        \State Let $c$ be an initial centroid estimate (e.g. the medoid of $O$) of length $n$
        \State Let $\tau$ and $\tau_0$ be two sequences of time stamps of length $n$ initialized with zero values
        \State Let $MeanK_0 = 0$ and $MeanK$ be two double values;  
        \Repeat
           \State $c_0 = c$, $\tau_0 = \tau$, $MeanK_0 = MeanK$;
           \State Evaluate $c$ and $\tau$ according to Eq. (\ref{eq:TEKA})
           \State //Average similarity between $c$ and $O$ elements
           \State $MeanK$=$\frac{1}{|O|}\sum_{o \in O} K(c,o)$
        \Until{$MeanK<MeanK_0$}
        \State ($c_0$, $\tau_0$) is the centroid estimation
        \State Uniformly re-sample $c_0$ using the time stamps $\tau_0$
      \end{algorithmic}
    \end{algorithm}

\begin{table*}[ht!]
\begin{center}
\begin{tabular}{cccc}

    %\textbf{iDBA}  & \textbf{CTW}& \textbf{TEKA} \\

    \textbf{DBA} &	\raisebox{-.5\height}{\includegraphics[scale=.22]{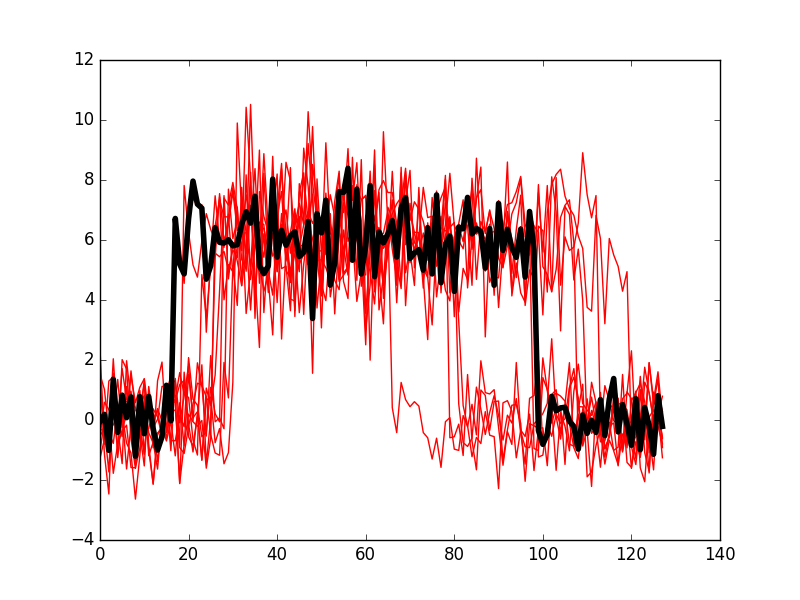}}&
    	\raisebox{-.5\height}{\includegraphics[scale=.22]{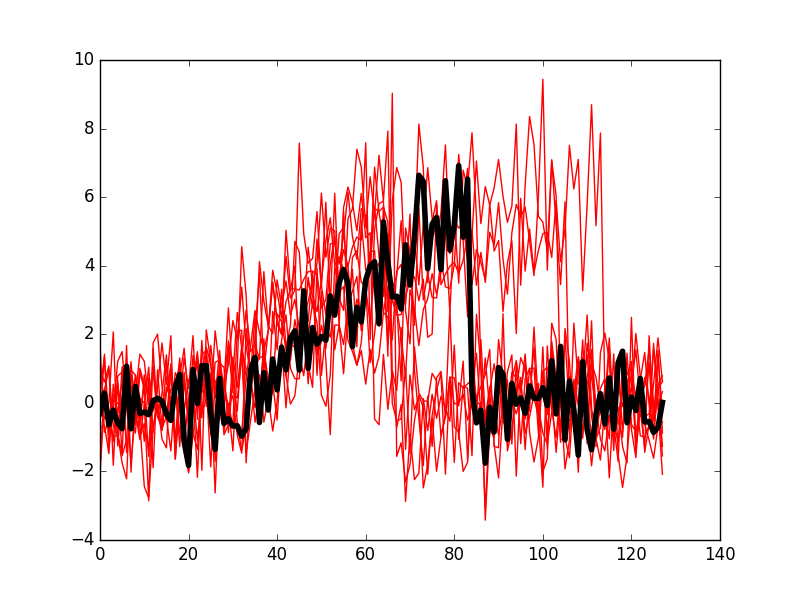}}&
    	\raisebox{-.5\height}{\includegraphics[scale=.22]{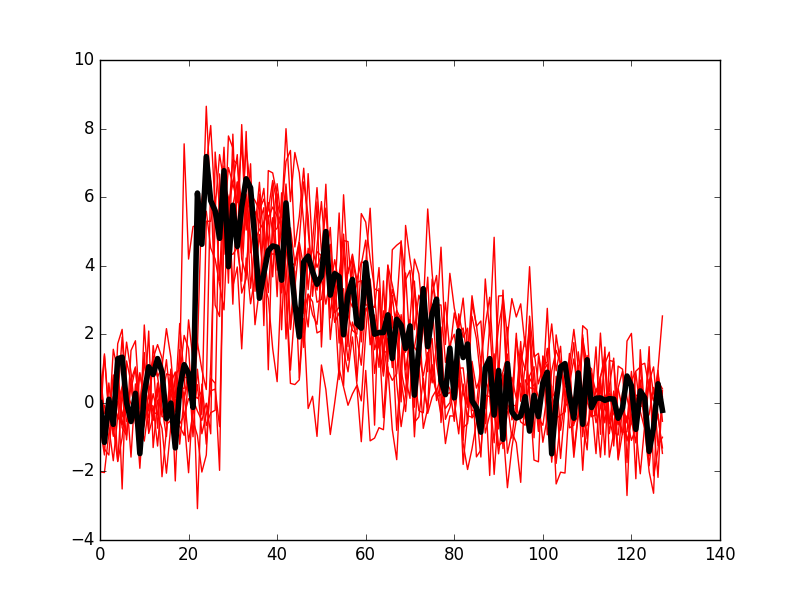}}\\
    \textbf{CTW} &	\raisebox{-.5\height}{\includegraphics[scale=.22]{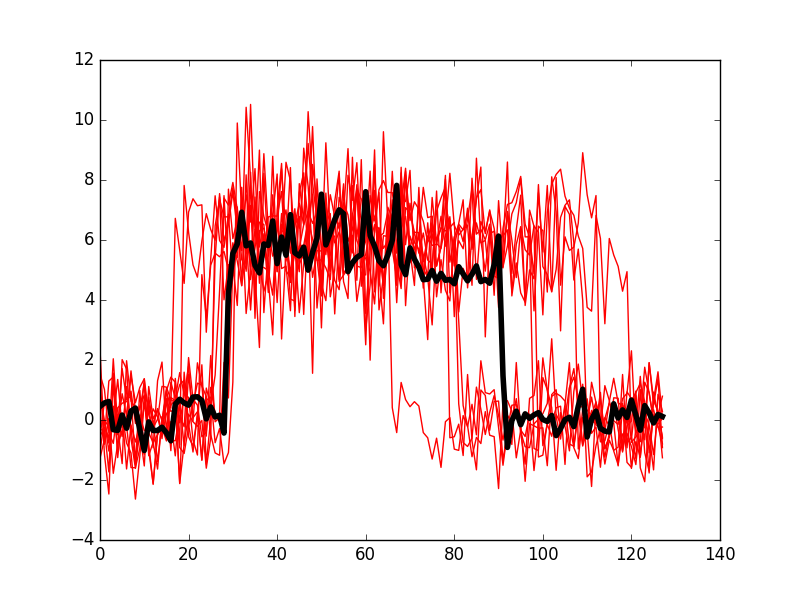}}&
    	\raisebox{-.5\height}{\includegraphics[scale=.22]{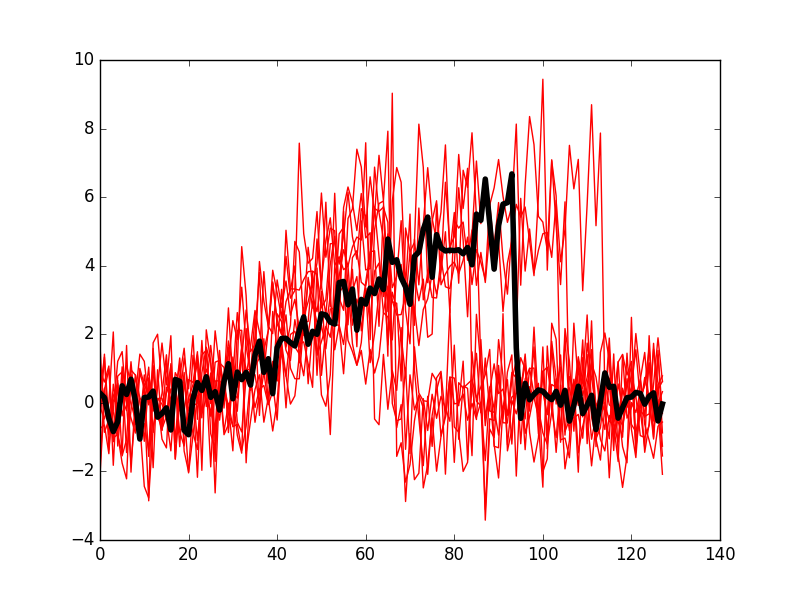}}&
    	\raisebox{-.5\height}{\includegraphics[scale=.22]{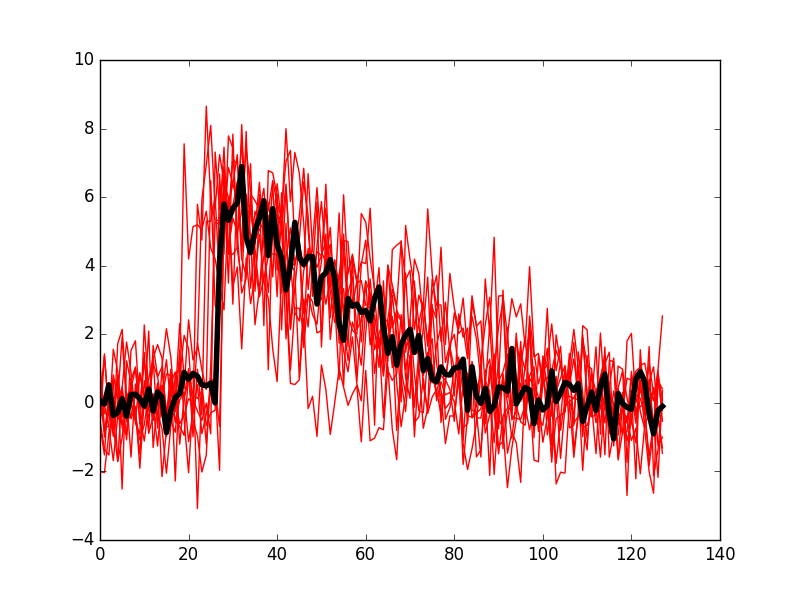}}\\
    \textbf{TEKA} &	\raisebox{-.5\height}{\includegraphics[scale=.22]{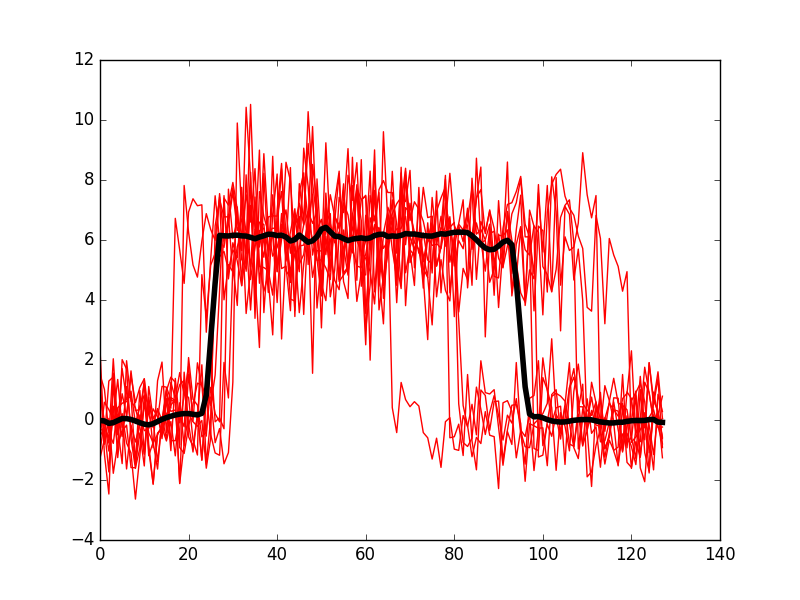}}&
    	\raisebox{-.5\height}{\includegraphics[scale=.22]{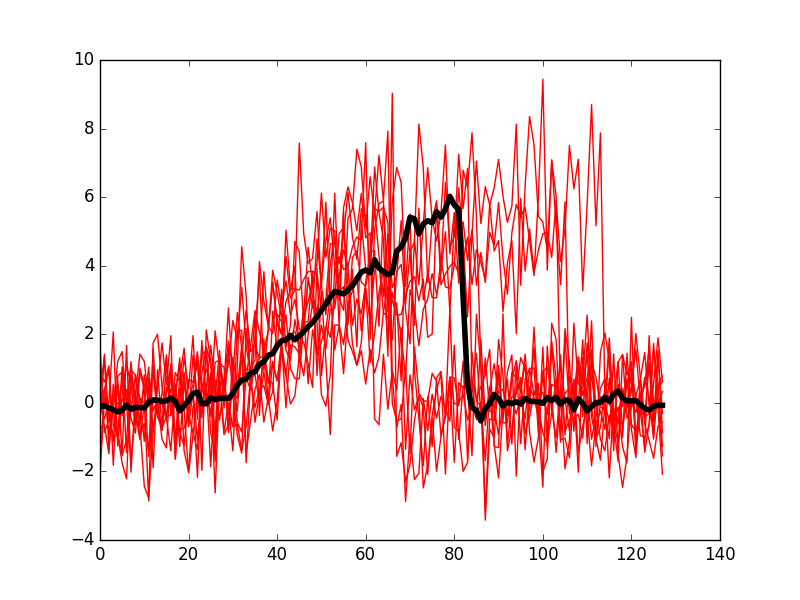}}&
    	\raisebox{-.5\height}{\includegraphics[scale=.22]{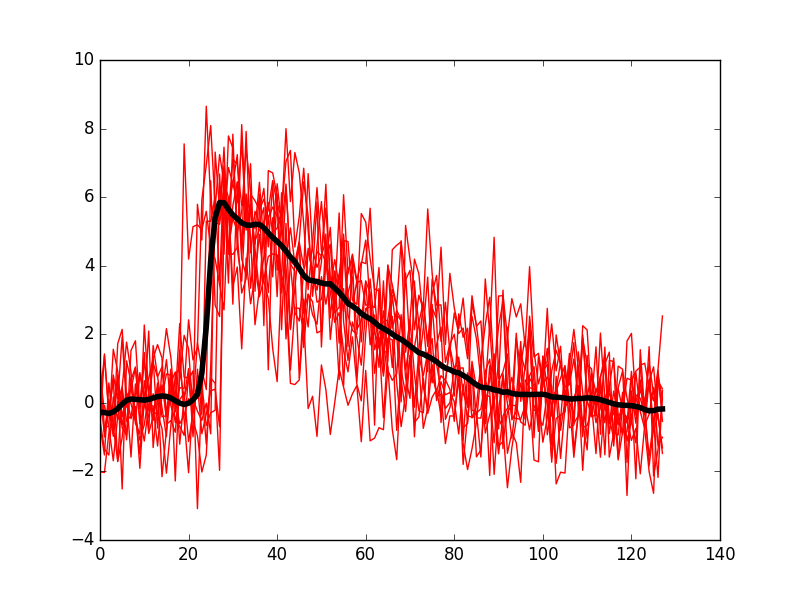}}\\

% %   	\includegraphics[scale=.3]{KDTW_centroid2_CBF_1.eps}&
%    	\includegraphics[scale=.22]{KDTW_centroid_CBF_1.eps}\\
%    \includegraphics[scale=.22]{DTW_centroid_CBF_2.eps}&
% %   \includegraphics[scale=.3]{KDTW_centroid2_CBF_2.eps}&
%    \includegraphics[scale=.22]{KDTW_centroid_CBF_2.eps}\\
%    	\includegraphics[scale=.22]{DTW_centroid_CBF_3.eps}&
%  %  	\includegraphics[scale=.3]{KDTW_centroid2_CBF_3.eps}&
%    	\includegraphics[scale=.22]{KDTW_centroid_CBF_3.eps}\\
%    	\hline 
    	\end{tabular}
\caption{Centroid estimation for the three categories of the CBF dataset and for the three tested algorithms: DBA (top), CTW  (center) TEKA (bottom). The centroid estimations are indicated as a bold black line superimposed on top of the time series (in light red) that are averaged.}
\label{fig:centroidsCBF}
\end{center}
\end{table*}

The figures presented in Table \ref{fig:centroidsCBF} compare the centroid estimates provided by the iterated DBA \cite{Petitjean2012}, CTW \cite{ZhouDLTore2009} and TEKA algorithms. For the experiment, the DBA and TEKA algorithms were iterated at most 10 times. The centroid estimates provided by the TEKA algorithm are much smoother than the ones provided by DBA or CTW. This denoising property, expected from any averaging algorithm, will be addressed in a dedicated experiment (c.f. subsection \ref{exp:denoising}).

\subsection{Role of parameter $\nu$}
In practice, the selectivity or bandwidth of the local alignment kernels (that is controlled by parameter $\nu$) has to be adapted according to the the lengths of the time series. If the time series are long, then $\nu$ should be reduced to maintain the calculability of the forward-backward matrices, and the local selectivity decreases. Hence, more alignment paths are likely and more sample pairs participate to the calculation of the average such that local details are filtered out by the averaging. Conversely if the time series are short, $\nu$ can be increased, hence fewer sample pairs participate to the calculation of the average, and details can be preserved.

\color{black}
\subsection{Computational complexity}
TEKA has intrinsically the same algorithmic complexity than the DBA algorithm, basically $O(L^2)$ for each pairwise averaging, where $L$ is the average length of the time series. Nevertheless, computationally speaking, TEKA algorithm is slightly  more costly mainly because of two reasons:
\begin{itemize}
\item the FB matrix induces a factor three in complexity because of the reverse alignment and the multiplication term by term of the forward and backward matrices.
\item the exponential terms that enter into the computation of KDTW (Eq. (\ref{Eq.KDTW})) are costly, basically $O(M(n) n^{1/2})$, where $M(n)$ is the cost of the floating point multiplication, and $n$ is the number of digits. This induces another factor 2 or 3 depending on the chosen floating point precision.
\end{itemize}

The overall algorithmic cost for averaging a set of $N$ time series of average length $L$ with an average number of iterations $I$ is, for the two algorithms, $O(I\cdot N \cdot L^2)$.\\

Some optimization are indeed possible, in particular replacing the exponential function by another local kernel easier to compute is an important source of algorithmic simplification. We do not address further this issue in this paper and let it stand as a perspective. 
 
\section{Experiments}
The two first proposed experiments aim at demonstrating the benefits of using time elastic centroids in a data reduction paradigm: 1-NC/NM (first near centroid or medoid) classification for the first one, and isolated gesture recognition for the second one using 1-NC/NM and SVM classifiers in conjunction with the KDTW kernel. The third experiment explores the noise reduction angle brought by time elastic centroids. 

\subsection{1-Nearest Centroid/Medoid classification}

%\color{red}
The purpose of this experiment is to evaluate the effectiveness of the proposed time elastic averaging method (TEKA) against a triple baseline. The first baseline allow us to compare centroid-based with medoid-based approaches. The second and third baselines are provided by the DBA \cite{Petitjean2012} and CTW \cite{ZhouDLTore2009} algorithms (thanks to the implementation proposed by the authors), currently considered as state of the art methods to average a set of sequences consistently with DTW. We have tested the CTW averaging with a 1-NC-DTW (CTW1) and a 1-NC-KDTW (CTW2) classifier to highlight the impact of the selected similarity measure. 

\color{black}
For this purpose, we empirically evaluate the effectiveness of the methods using a first nearest centroid/medoid (1-NC/NM) classification task on a set of time series derived from widely diverse fields of application. The task consists of representing each category contained in a training data set by estimating its medoid or centroid and then evaluating the error rate of a 1-NC classifier on an independent testing data set. Hence, the classification rule consists of assigning to the tested time series the category which corresponds to the closest (or most similar) medoid or centroid according to the DTW measure for DTW medoid (DTW-M), DBA and CTW centroids (CTW1) or to KDTW measure for KDTW medoid (KDTW-M), CTW (CTW2) and TEKA centroids. \\

In \cite{Petitjean2014} a generalized k-NC task is described. The authors demonstrate that by selecting the appropriate number $k$ of centroids (using DBA and k-means), they achieve, without loss, a 70\% speed-up in average, compared to the original k-Nearest Neighbor task. Although, in general, the classification accuracy is improved when several centroids are used to represent the training datasets,  our main purpose is to highlight and amplify the  discrimination between time series averaging methods: this is why we stick here with the 1-NC task. 

%DBA iterative centroid methods is iterated at most 20 times and yield local estimates of the centroid. The pKDTW-PWA progressive agglomerative centroid method is only processed once, and hence is roughly 10 times faster than DBA.\\

\begin{table*}[]
\caption{Comparative study using the UCR and UCI data sets: classification error rates evaluated on the TEST data set (in \%) obtained using the first nearest neighbour classification rule for DTW-M,  KDTW-M,  (medoids), DBA, CTW1, CTW2 and TEKA (centroids). A single medoid/centroid extracted from the training data set represents each category.}
\label{tab:classResults}
\centering
\resizebox{\textwidth}{!}{\begin{tabular}{|l|c|c|c|c|c|c|c|c|}
\hline
\textbf{DATASET} &  \# Cat $|$ L &   \textbf{DTW-M} &  \textbf{DBA} &  \textbf{CTW1} &  \textbf{CTW2} &  \textbf{ KDTW-M}  &  \textbf{TEKA} \\
\hline\hline
Synthetic\_Control	&	6$|$60	&	3.00	&	\textbf{2.00}	&	19.00	&	3.33	&	3.33	&	2.33	\\
Gun\_Point	&	2$|$150	&	44.00	&	32.00	&	54.67	&	\textbf{25.33}	&	52.00	&	27.33	\\
CBF	&	3$|$128	&	7.89	&	5.33	&	34.22	&	3.55	&	8.11	&	\textbf{3.33}	\\
Face\_(all)	&	14$|$131	&	25.21	&	18.05	&	34.38	&	27.93	&	20.53	&	\textbf{13.61}	\\
OSU\_Leaf	&	6$|$427	&	64.05	&	56.20	&	64.05	&	57.02	&	53.31	&	\textbf{50.82}	\\
Swedish\_Leaf	&	15$|$128	&	38.56	&	30.08	&	32	&	25.76	&	31.36	&	\textbf{22.08}	\\
50Words	&	50$|$270	&	48.13	&	41.32	&	48.57	&	36.48	&	23.40	&	\textbf{19.78}	\\
Trace	&	4$|$275	&	\textbf{5.00}	&	7.00	&	6.00	&	18	&	23.00	&	16.00	\\
Two\_Patterns	&	4$|$128	&	1.83	&	1.18	&	26.75	&	37.75	&	1.17	&	\textbf{1.10}	\\
Wafer	&	2$|$152	&	64.23	&	33.89	&	37.83	&	33.27	&	43.92	&	\textbf{8.38}	\\
Face\_(four)	&	4$|$350	&	12.50	&	13.64	&	19.32	&	15.91	&	17.05	&	\textbf{10.23}	\\
Lightning-2	&	2$|$637	&	34.43	&	37.70	&	37.70	&	\textbf{29.51}	&	\textbf{29.51}	&	\textbf{29.51}	\\
Lightning-7	&	7$|$319	&	27.40	&	27.40	&	41.10	&	38.35	&	19.18	&	\textbf{16.44}	\\
ECG200	&	2$|$96	&	32.00	&	28.00	&	27.00	&	\textbf{25}	&	29.00	&	26.00	\\
Adiac	&	37$|$176	&	57.54	&	52.69	&	54.73	&	34.78	&	40.67	&	\textbf{32.22}	\\
Yoga	&	2$|$426	&	47.67	&	47.87	&	53.56	&	48.97	&	47.53	&	\textbf{44.90}	\\
Fish	&	7$|$463	&	38.86	&	30.29	&	39.42	&	22.28	&	20.57	&	\textbf{14.28}	\\
Beef	&	5$|$470	&	60.00	&	53.33	&	53.33	&	\textbf{50}	&	53.33	&	\textbf{50}	\\
Coffee	&	2$|$286	&	57.14	&	32.14	&	32.14	&	\textbf{28.57}	&	32.14	&	32.14	\\
OliveOil	&	4$|$570	&	26.67	&	\textbf{16.67}	&	13.33	&	23.33	&	30	&	\textbf{16.67}	\\
CinC\_ECG\_torso	&	4$|$1639	&	74.71	&	53.55	&	73.33	&	42.90	&	66.67	&	\textbf{33.04}	\\
ChlorineConcentration	&	3$|$166	&	65.96	&	68.15	&	67.40	&	67.97	&	65.65	&	\textbf{64.97}	\\
DiatomSizeReduction	&	4$|$345	&	22.88	&	5.88	&	5.23	&	\textbf{2.61}	&	11.11	&	2.94	\\
ECGFiveDays	&	2$|$136	&	47.50	&	30.20	&	34.49	&	13.47	&	\textbf{11.38}	&	16.37	\\
FacesUCR	&	14$|$131	&	27.95	&	18.44	&	32.20	&	21.66	&	20.73	&	\textbf{12.19}	\\
Haptics	&	5$|$1092	&	68.18	&	64.61	&	58.77	&	57.47	&	63.64	&	\textbf{53.57}	\\
InlineSkate	&	7$|$1882	&	78.55	&	76.55	&	81.64	&	82.18	&	78.36	&	\textbf{75.09}	\\
ItalyPowerDemand	&	2$|$24	&	31.68	&	20.99	&	15.84	&	9.33	&	\textbf{5.05}	&	6.61	\\
MALLAT	&	8$|$1024	&	6.95	&	6.10	&	5.24	&	\textbf{3.33}	&	6.87	&	3.66	\\
MedicalImages	&	10$|$99	&	67.76	&	58.42	&	58.29	&	59.34	&	\textbf{57.24}	&	59.60	\\
MoteStrain	&	2$|$84	&	15.10	&	13.18	&	19.01	&	15.33	&	12.70	&	\textbf{9.35}	\\
SonyAIBORobot\_SurfaceII	&	2$|$65	&	26.34	&	21.09	&	20.57	&	\textbf{17.52}	&	26.230	&	19.30	\\
SonyAIBORobot\_Surface	&	2$|$70	&	38.10	&	19.47	&	14.48	&	\textbf{9.31}	&	39.77	&	17.95	\\
Symbols	&	6$|$398	&	7.64	&	4.42	&	22.31	&	20.70	&	\textbf{3.92}	&	4.02	\\
TwoLeadECG	&	2$|$82	&	24.14	&	\textbf{13.17}	&	20.37	&	19.23	&	27.04	&	18.96	\\
WordsSynonyms	&	25$|$270	&	70.85	&	64.26	&	78.84	&	63.32	&	64.26	&	\textbf{56.11}	\\
Cricket\_X	&	12$|$300	&	67.69	&	\textbf{52.82}	&	78.46	&	73.85	&	61.79	&	\textbf{52.82}	\\
Cricket\_Y	&	12$|$300	&	68.97	&	52.82	&	69.74	&	65.64	&	\textbf{46.92}	&	50.25	\\
Cricket\_Z	&	12$|$300	&	73.59	&	\textbf{48.97}	&	78.21	&	64.36	&	56.67	&	51.79	\\
uWaveGestureLibrary\_X	&	8$|$315	&	38.97	&	33.08	&	37.33	&	34.61	&	34.34	&	\textbf{32.18}	\\
uWaveGestureLibrary\_Y	&	8$|$315	&	49.30	&	44.44	&	45.42	&	41.99	&	42.18	&	\textbf{39.64}	\\
uWaveGestureLibrary\_Z	&	8$|$315	&	47.40	&	\textbf{39.25}	&	47.65	&	39.36	&	41.96	&	39.97	\\
PWM2	&	3$|$128	&	43.00	&	35.00	&	63.66	&	6.33	&	21.00	&	\textbf{4.33}	\\
uWaveGestureLibrary\_3D	&	8$|$315	&	10.11	&	\textbf{5.61}	&	9.35	&	7.68	&	13.74	&	7.73	\\
CharTrajTT\_3D	&	20$|$178	&	11.026	&	9.58	&	13.45	&	15.05	&	6.93	&	\textbf{4.99}	\\
\hline
\hline
\textbf{\# Best Scores} &  - &  1 &   7 & 0 & 9 & 6 &   \textbf{27}   \\
\hline
\textbf{\# Uniquely Best Scores} &  - &  1 &  5 & 0 & 7 &  5 &   \textbf{23}   \\
\hline
\textbf{Average rank} &  - &  4.56 &  2.87 & 4.62 & 2.97 &  3.22  &   \textbf{1.6} \\
\hline
\end{tabular}}
\end{table*} 

%\FloatBarrier

%\begin{figure*}[h!]
%\centering
%\begin{tabular}{cc}
%	%\includegraphics[scale=.4, angle=0]{FriedmanTestSC.eps} 
%	%&
%   	\includegraphics[scale=.5, angle=0]{FriedmanTestBP.eps} 
%\end{tabular}
%\caption{\textit{Post hoc} analysis of the Friedman's test: ($A_1$) DTW$_{Medoid}$, ($A_2$) DBA, ($A_3$) KDTW$_{Medoid}$ and ($A_4$) TEKA.}
%\label{fig:FriedmanTest}
%\end{figure*}

A collection of 45 heterogeneous data sets is used to assess the proposed algorithms. The collection includes synthetic and real data sets, as well as univariate and multivariate time series. These data sets are distributed as follows: \\
\begin{itemize}
\item 42 of these data sets are available at the UCR repository \cite{KeoghUCRdataset}. Basically, we used all the data sets except for \textit{StarLightCurves}, \textit{Non-Invasive Fetal ECG Thorax1} and  \textit{Non-Invasive Fetal ECG Thorax2}. Although these last three data sets are still tractable, their computational cost is high because of their size and the length of the time series they contain. All these data sets are composed of scalar time series.
\item One data set, uWaveGestureLibrary\_3D was constructed from the uWaveGestureLibrary\_{X|Y|Z} scalar data sets to compose a new set of multivariate (3D) time series.
\item One data set, CharTrajTT, is available at the UCI Repository \cite{Lichman:2013} under the name \textit{Character Trajectories Data Set}. This data set contains multivariate (3D) time series and is divided into two equal sized data sets (TRAIN and TEST) for the experiment.  
\item The last data set, \textit{PWM2}, which stands for Pulse Width Modulation \cite{PWM}, was specifically defined to demonstrate a weakness in dynamic time warping (DTW) pseudo distance. This data set is composed of synthetic scalar time series.\\
\end{itemize}

For each dataset, a training subset (TRAIN) is defined as well as an independent testing subset (TEST). We use the training sets to extract single medoids or centroid estimates for each of the categories defined in the data sets. 

Furthermore, for KDTW-M, CTW2  and TEKA, the $\nu$ parameter is optimized using a \textit{leave-one-out} (LOO) procedure carried out on the TRAIN data sets. The $\nu$ value is selected within the discrete set $\{.01, .05, .1, .25, .5, .75, 1, 2, 5, 10, 15, 20, 25, 50, 100\}$. The value that minimizes the LOO classification error rate on the TRAIN data is then used to provide the error rates that are estimated on the TEST data.\\

%\color{red}
The classification results are given in Table \ref{tab:classResults}. It can be seen from this experiment, that 
\begin{enumerate}[i)]
\item Centroid-based methods outperform medoid-based methods: DBA and CTW (CTW2) yield lower error rates compared to DTW-M, as do TEKA compared to KDTW-M and DTW-M.
\item CTW pairs much better with KDTW (CTW2 outperforms CTW1)
\item TEKA outperforms DBA (under the same experimental conditions (maximum of 10 iterations)), and CTW.\\ 
\end{enumerate}

The average ranking for all six tested methods, which supports our preliminary conclusion, is given at the bottom of Table \ref{tab:classResults}.\\

\begin{table}[h!]
\centering
\caption{ Wilcoxon signed-rank test of pairwise accuracy differences for 1-NC/NM classifiers carried out on the 45 datasets. }
\resizebox{.48 \textwidth}{!}{\begin{tabular}{|l|l|l|l|l|l|}
\hline
Method	           & KDTW-M & DBA & CTW1 & CTW2 & TEKA\\ \hline
\hline 
 DTW-M & \textbf{\texttt{p<.0001}} & \textbf{\texttt{p<.0001}} & \texttt{0.638} & \textbf{\texttt{0.0002}} & \textbf{\texttt{p<.0001}}\\ \hline
 KDTW-M & -  & \texttt{0.395} & \textbf{\texttt{0.0004}} & \texttt{0.5261} & \textbf{\texttt{p<.0001}}\\ \hline
 DBA     &  -      & -  & \textbf{\texttt{p<.0001}} & \texttt{0.8214} & \textbf{\texttt{p<.0001}}\\ \hline 
 CTW1     & - & -  & - & \textbf{\texttt{p<.0001}} & \textbf{\texttt{p<.0001}}\\ \hline 
 CTW2     & - & -  & - & - & \textbf{\texttt{p<.0001}}\\ \hline 
\end{tabular}}
\label{tab:significance_test}
\end{table}

In Table \ref{tab:significance_test} we report the P-values for each pair of tested algorithms using a Wilcoxon  signed-rank test. The null hypothesis is that for a tested pair of classifiers, the difference between classification error rates obtained on the 45 datasets follows a symmetric distribution around zero. With a $.05$ significance level, the P-values that lead to reject the null hypothesis are shown in bolded fonts in the table.
 This analysis confirms our previous analysis of the classification results. We observe that  centroid-based approaches perform significantly better than medoid-based approaches.  Furthermore, KDTW-M appears to be significantly better than DTW-M. 

Furthermore, TEKA is evaluated as significantly better than DBA and CTW2 in this experiment. Note also that DBA does not seem to perform significantly better than KDTW-M or CTW2, and that CTW1 performed similarly to DTW-M and poorly compared to the other centroid methods. Hence, it confirms out that CTW method seems to pair well with KDTW measure but poorly with the DTW measure.

\color{black}

\subsection{Instance set reduction}

In this second experiment, we address an application that consists in summarizing subsets of training time series to speed-up an isolated gesture recognition process.

The dataset that we consider enables to explore the hand-shape and the upper body movement using 3D positions of skeletal joints captured using a Microsoft Kinect 2 sensor. 20 subjects have been selected (15 males and 5 females) to perform in front of the sensor (at a three meters distance) the six selected NATOPS gestures. Each subject repeated each gesture three times. Hence the isolated gesture dataset is composed of 360 gesture utterances that have been manually segmented to a fixed length of 51 frames \footnote{These datasets will be made available for the community at the earliest feasible opportunity}. %(1.7 sec. duration). The same 20 subjects also performed 3 times these gestures in a continuous mode. The obtained dataset consists of 60 samples of motion data stream, each stream containing one occurrence of each gesture.

%\color{red}
To evaluate this task, we have performed a subject cross validation experiment consisting of 100 tests: for each test, 10 subjects have been randomly drawn among 20 for training and the remaining 10 subjects have been retained for testing. 1-NN/NC (our baselines) and SVM classifiers are evaluated, with or without summarizing the subsets composed with the three repetitions performed by each subjects using a single centroid (DBA, CTW, TEKA) or Medoid (KDTW-M). The $\nu$ parameter of the KDTW kernel as well as the SVM meta parameter (RBF bandwidth $\sigma$ and $C$) are optimized using a leave one subject procedure on the training dataset. The kernels $exp(-DTW(.,.)/\sigma)$ and $exp(-KDTW(.,.)/\sigma)$ are used respectively in the SVM DTW and SVM KDTW classifiers.   \\

%\color{red}
\begin{table}[h!]
\centering
\caption{Assessment measures (ERR:Error rate, PRE: Precision, REC:Recall and $\mathbf{F_1}$ score) for the  isolated gestures recognition. $\overline{\#Ref}$ is the number of training gestures for the 1-NN/NC classifiers and the mean number of support vectors for the SVM classifiers.}
\resizebox{.48 \textwidth}{!}{\begin{tabular}{|l|l|l|l|l|l|}
\hline
\textbf{Method} & \begin{tabular}[c]{@{}l@{}}\textbf{ERR}\\ mean $\|$ std\end{tabular} & \textbf{PRE} & \textbf{REC} & $\mathbf{F_1}$ & \textbf{$\overline{\#Ref}$}\\ \hline \hline
1-NN DTW         & .134 $\|$ .012 & .869 & .866 & 0.867 & 180\\ \hline
1-NN KDTW        & \textbf{.128} $\|$ .016 & .876 & .972 & .874 & 180\\ \hline
\hline
1-NC DTW-DBA    &  .136 $\|$  .014  &  .868   & .864  & .866 & \textbf{60}\\ \hline
1-NC KDTW-CTW    &   .135 $\|$ .016 & .871    & .865  & .868 & \textbf{60}\\ \hline
1-NC KDTW-TEKA      & \textbf{.133} $\|$ .014 & .871 & .867 & .869 & \textbf{60} \\ \hline 
\hline
SVM DTW         & .146 $\|$ .015 & .871 & .854 & .862 & 164.97
 \\ \hline
SVM KDTW        & \textbf{.051} $\|$ .015 & .952 & .949 & .951 & \textbf{103.10}
\\ \hline \hline
SVM KDTW-M        & .087 $\|$ .02 & .92.9 & .92.6 & .92.7 & 47.62
\\ \hline
SVM KDTW-DBA      & .080 $\|$ .017 & .935 & .931 & .931 & \textbf{46.74}\\ \hline
SVM KDTW-CTW      & .085 $\|$ .021 & .933 & .927 & .930 & 50.12\\ \hline
SVM KDTW-TEKA      & \textbf{.079} $\|$ .019 & .937 & .933 & .935 & 47.45\\ \hline
\end{tabular}}
\label{tab:isol_res}
\end{table}

\begin{table}[h!]
\centering
\caption{ Wilcoxon signed-rank test of pairwise accuracy differences for 1-NN/NC classifiers. DTW and KDTW methods exploit the entire training sets while  the other methods only use one centroid for each subject and each gesture label.}
\resizebox{.48 \textwidth}{!}{\begin{tabular}{|l|l|l|l|l|}
\hline
Method          & 1-NN & 1-NC & 1-NC & 1-NC\\ 
                & KDTW & DBA & CTW & TEKA\\ \hline
\hline 
 1-NN DTW     & \textbf{\texttt{p<.0001}} & \texttt{0.140} & \texttt{0.886} & \texttt{0.371} \\ \hline
 1-NN KDTW    & -       & \textbf{\texttt{p<.0001}} & \textbf{0.026} &  \texttt{0.087}\\ \hline
 1-NC DBA    &  -      & -  & \texttt{0.281} &  \textbf{\texttt{0.006}} \\ \hline 
 1-NC CTW     & - & -  & - & \texttt{0.199} \\ \hline 
\end{tabular}}
\label{tab:significance_1NN}
\end{table}

\begin{table}[h!]
\centering
\caption{ Wilcoxon signed-rank test oof pairwise accuracy differences for SVM classifiers. DTW and KDTW methods exploit the entire training sets while  the other methods only use  one centroid for each subject and each gesture label.}
\resizebox{.48 \textwidth}{!}{\begin{tabular}{|l|l|l|l|l|l|}
\hline
Method          & SVM  & SVM    & SVM & SVM & SVM \\ 
	            & KDTW & KDTW-M & DBA & CTW & TEKA\\ \hline
\hline 
 SVM DTW     & \textbf{\texttt{p<.0001}} & \textbf{\texttt{p<.0001}} & \textbf{\texttt{p<.0001}} & \textbf{\texttt{p<.0001}} & \textbf{\texttt{p<.0001}} \\ \hline
 SVM KDTW    & -       &  \textbf{\texttt{p<.0001}} &\textbf{\texttt{p<.0001}} & \textbf{\texttt{p<.0001}} &  \textbf{\texttt{p<.0001}}\\ \hline
 SVM KDTW-M    & -   &  -  & \textbf{\texttt{0.002}} & \texttt{0.57} &  \textbf{\texttt{0.0002}}\\ \hline
 SVM DBA    &  -   & -  & -  & \texttt{0.107} &  \texttt{0.339} \\ \hline 
 SVM CTW    & - & - & -  & - & \textbf{\texttt{0.013}} \\ \hline 
\end{tabular}}
\label{tab:significance_SVM}
\end{table}

Table \ref{tab:isol_res} gives the assessment measures (ERR: average error rate, PRE: macro average precision, REC: macro average recall and $F_1 = 2 \cdot \frac{\mathrm{precision} \cdot \mathrm{recall}}{\mathrm{precision} + \mathrm{recall}}$) for the isolated gestures classification task. In addition, the number of reference instances used by the 1-NN/NC classifiers or the number of support vectors exploited by the SVM ($\overline{\#Ref}$ column in the table) are reported to demonstrate the data reduction that is induced by the methods in the training sets.

The results show that the DTW measure does not fit well with SVM comparatively to KDTW: the error rate or the F1 score are about $9\%$ higher or lower for the isolated gesture task. Hence, to compare the DBA, CTW and TEKA centroids using a SVM classification, the KDTW kernel has been used.  When using the centroids (SVM KDTW-DBA, SVM KDTW-CTW, SVM KDTW-TEKA), or Medoids (SVM KDTW-M) the error rate or $F_1$ score increases or decreases only by around $2.5\%$ and $2\%$ comparatively to the SVM-KDTW that achieves the best scores. Meanwhile the number of support vectors exploited by the SVM drops by a two factor, leading to an expected speed-up of $2$. Compared to 1-NN classification without centroids, the SVM KDTW with centroids achieves a much better performance, with an  expected speed-up of $4$ ($\sim 50$ support vectors comparatively to $180$ gesture instances). This demonstrates the capacity of centroid methods to reduce significantly the size of the training sets while maintaining a very similar level of accuracy.

In more details, the TEKA is the centroid-based method that achieves the lowest error rates for the two classification tasks, while DBA is the centroid-based method that exploits the fewest support vectors (46.5). 
 
Table \ref{tab:significance_1NN} and \ref{tab:significance_SVM} give the P-values for the Wilcoxon  signed-rank tests.  With the same null hypothesis as above (difference between the error rates follows a symmetric distribution around zero), and with a $.05$ significance level, the P-values that lead to reject the null hypothesis are presented in bolded fonts in the tables. From Table \ref{tab:significance_1NN} we note that 1NN-KDTW (which exploits the full training set) performs significantly better than 1NN DTW, 1-NC DTW-DBA and 1-NC KDTW-CTW but not significantly than 1-NC KDTW-TEKA. Conversely, 1-NC KDTW-TEKA performs significantly better that 1-NC DTW-DBA but not significantly better that 1-NC KDTW-CTW.
Similarly, from Table \ref{tab:significance_SVM} we observe that SVM KDTW, which exploits the full training set, performs significantly better than all centroid or medoid based methods. Also, SVM KDTW-TEKA performs significantly better than SVM KDTW-CTW but not significantly better than SVM KDTW-DBA. Finally SVM KDTW-TEKA and SVM KDTW-DBA outperform the medoid based method (SVM KDTW-M) but not SVM KDTW-CTW.

If the three centroid methods show rather close accuracies on this experiment, TEKA is significantly better than DBA on the 1NC task and significantly better than CTW on the SVM task.

\color{black}
\subsection{Denoising experiment}
\label{exp:denoising}
To demonstrate the utility of centroid based methods for denoising data, we construct a demonstrative synthetic experiment that provides some insights. The test is based on the following 2D periodic signal:

\begin{align}
X_k(t)=\left(A_k+B_k\sum_{i=1}^\infty \delta(t-\frac{2\pi i}{6\omega_k})\right) cos(\omega_k t+\phi_k)\\
Y_k(t)=\left(A_k+B_k\sum_{i=1}^\infty \delta(t-\frac{2\pi i}{6\omega_k})\right) sin(\omega_k t+\phi_k)\nonumber
\end{align}
where $A_k=A_0+a_k$, $B_k=(A_0+5)+b_k$ and $\omega_k =  \omega_0 + w_k$,  $A_0$ and $\omega_0$  are constant and $a_k$, $b_k$, $\omega_k$, $\phi_k$ are small perturbation in amplitude, frequency and phase respectively and randomly drawn from $a_k \in [0, A_0/10]$, $b_k \in [0, A_0/10]$, $\omega_k \in [-\omega_0/6.67, \omega_0/6.67]$, $\phi_k \in [-\omega_0/10, \omega_0/10]$.

\begin{figure}[]
\centering
\begin{tabular}{ccc}
	\includegraphics[scale=.35, angle=0]{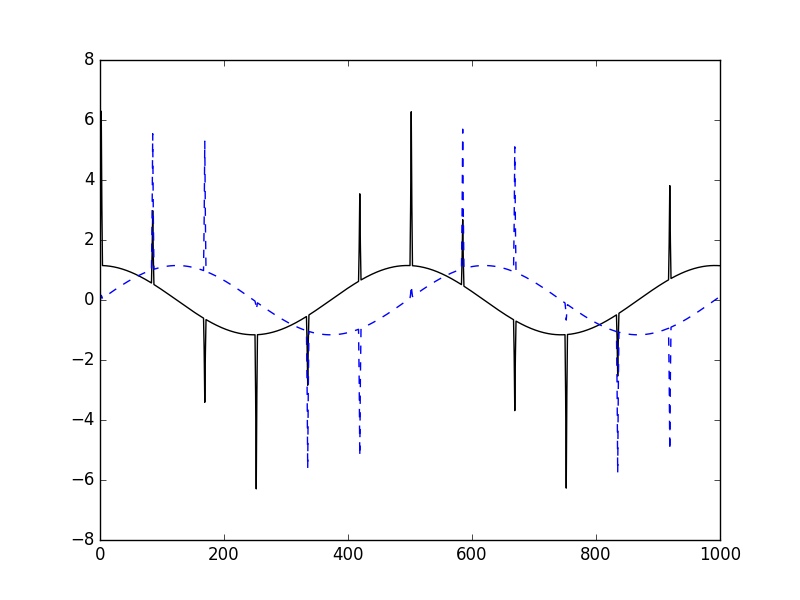} \\
    \includegraphics[scale=.35, angle=0]{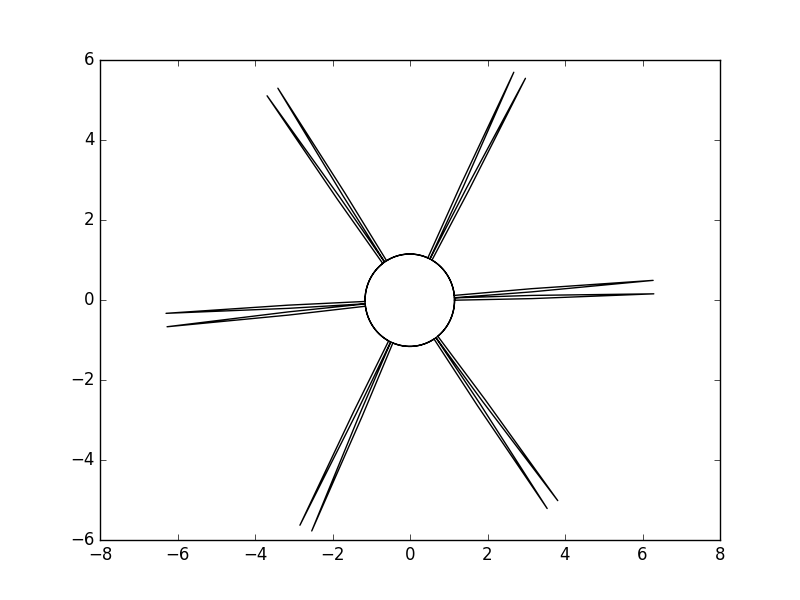} 
\end{tabular}
\caption{$(\tilde{X_k}(t),\tilde{Y_k}(t))$ waveforms (top) and corresponding 2D shape (bottom, plain black curve) of the synthetic signal.}
\label{fig:cleanShape}
\end{figure}

\begin{figure}[]
\centering
\includegraphics[scale=.35, angle=0]{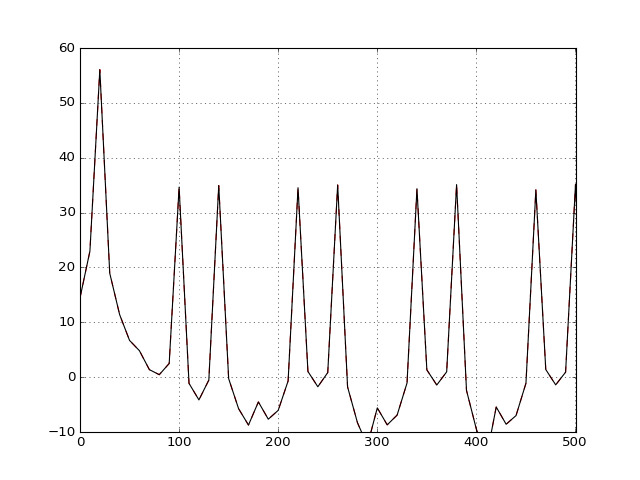}
\caption{Log power spectra of a $\tilde{X_k}$ component.}
\label{fig:cleanSpectra}
\end{figure}

\begin{figure}[]
\centering
\begin{tabular}{cc}
\includegraphics[scale=.35, angle=0]{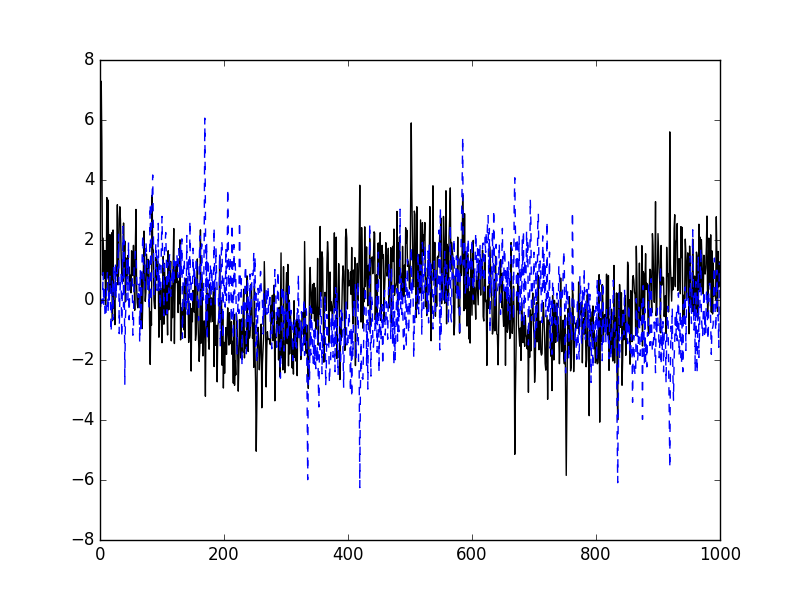}\\
\includegraphics[scale=.35, angle=0]{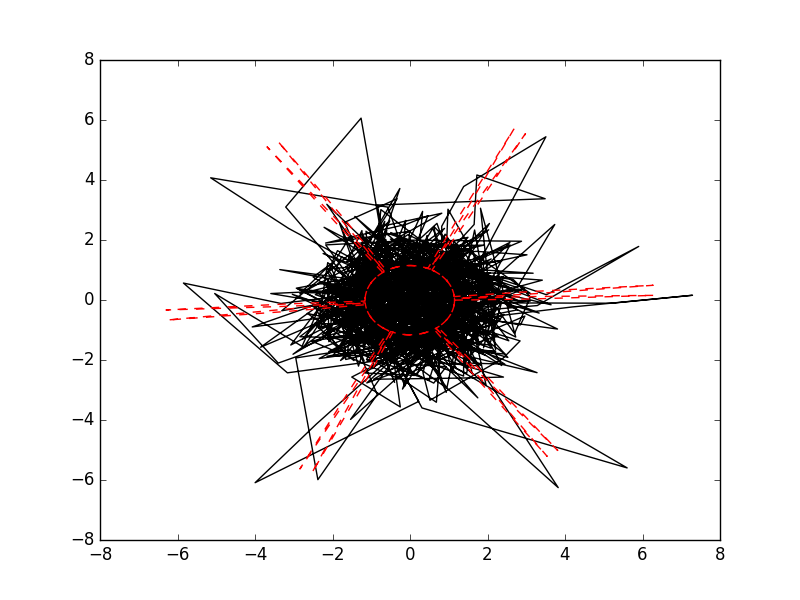}
\end{tabular}
\caption{Noisy $(x_k(t),y_k(t))$ waveforms (top) and corresponding 2D shape (bottom) of the synthetic signal.}
\label{fig:noisyShape}
\end{figure}

%\begin{figure*}[ht]
%\centering
%\begin{tabular}{cccc}
%\includegraphics[scale=.2, angle=0]{starSig_fft_Ceuclid.png} &
%\includegraphics[scale=.2, angle=0]{starSig_fft_DBA.png} &
%\includegraphics[scale=.2, angle=0]{starSig_fft_ctw.png} &
%\includegraphics[scale=.2, angle=0]{starSig_fft_TEKA.png} 
%\end{tabular}
%\caption{Log power spectra for the DBA centroid (left, black bold line) and the TEKA centroid (right, black bold line).  The spectra for the noisy signal and the noise free signal are given respectively in red '+' mark  and black dotted lines.}
%\label{fig:spectraCentroids}
%\end{figure*}

\begin{figure*}[h]
\centering
\begin{tabular}{cccc}
Euclidean & DBA & CTW & TEKA \\
\includegraphics[scale=.2, angle=0]{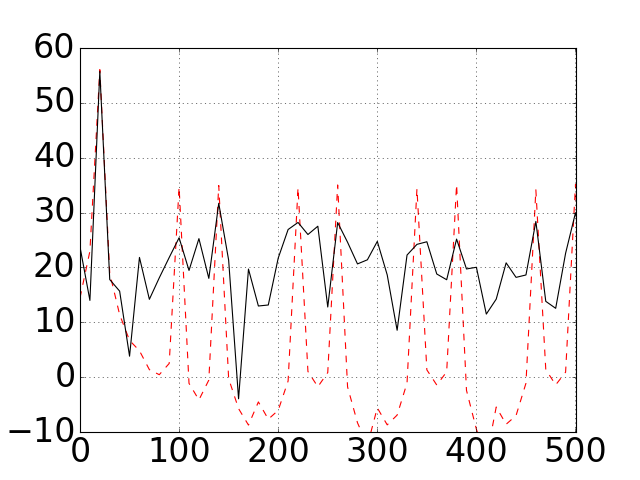} &
\includegraphics[scale=.2, angle=0]{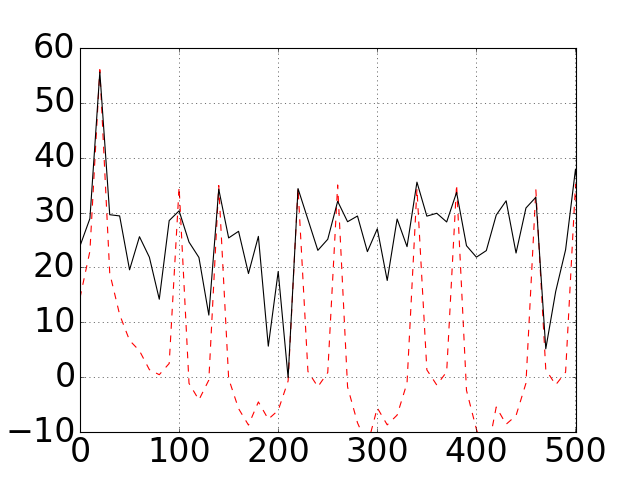} &
\includegraphics[scale=.2, angle=0]{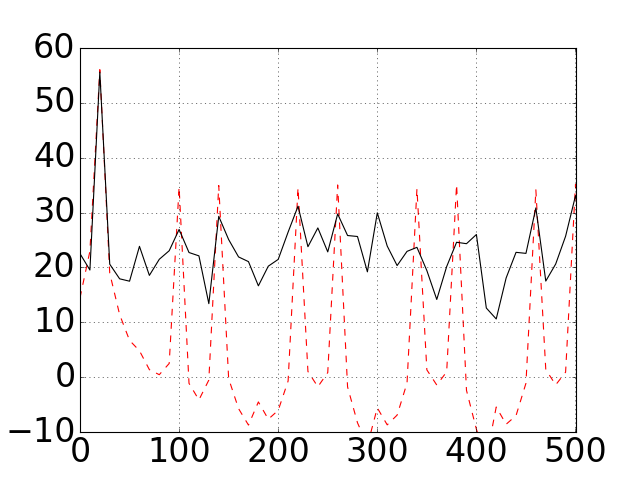} &
\includegraphics[scale=.2, angle=0]{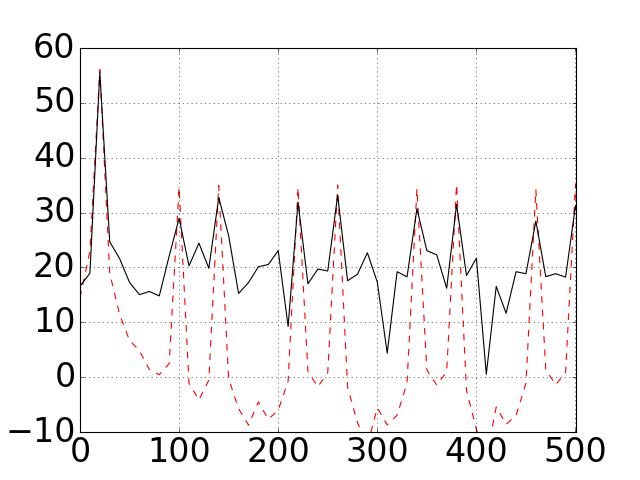} \\

\includegraphics[scale=.2, angle=0]{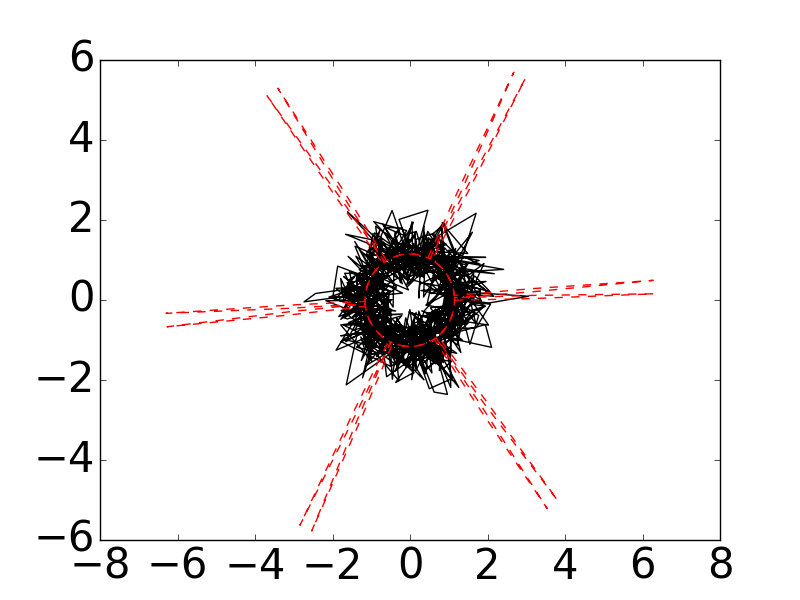} &
\includegraphics[scale=.2, angle=0]{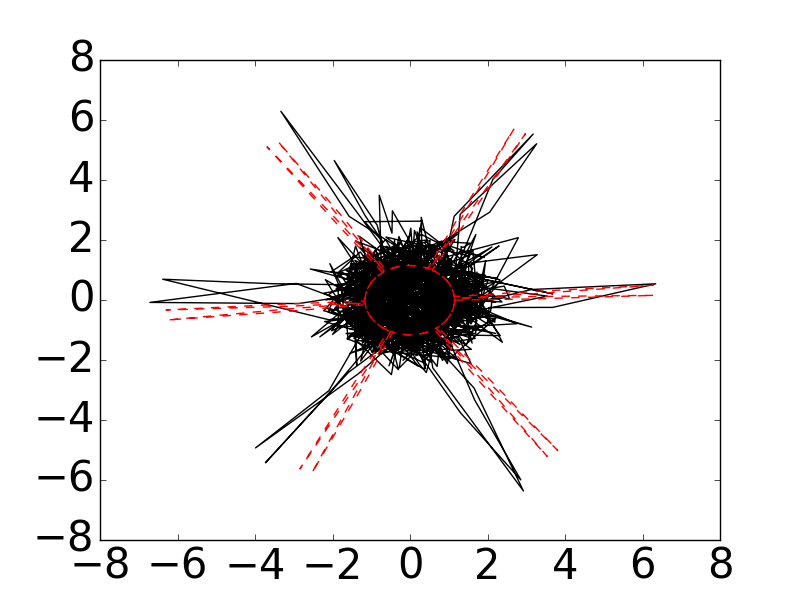} &
\includegraphics[scale=.2, angle=0]{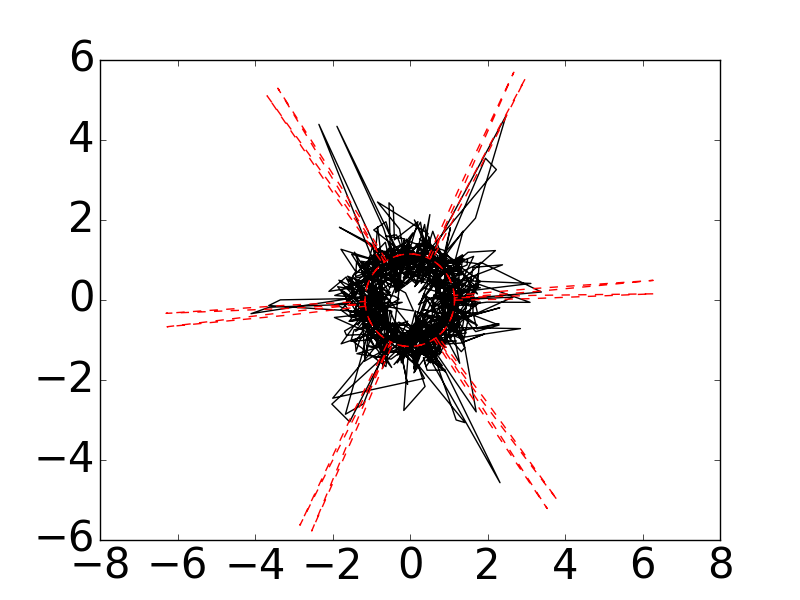} &
\includegraphics[scale=.2, angle=0]{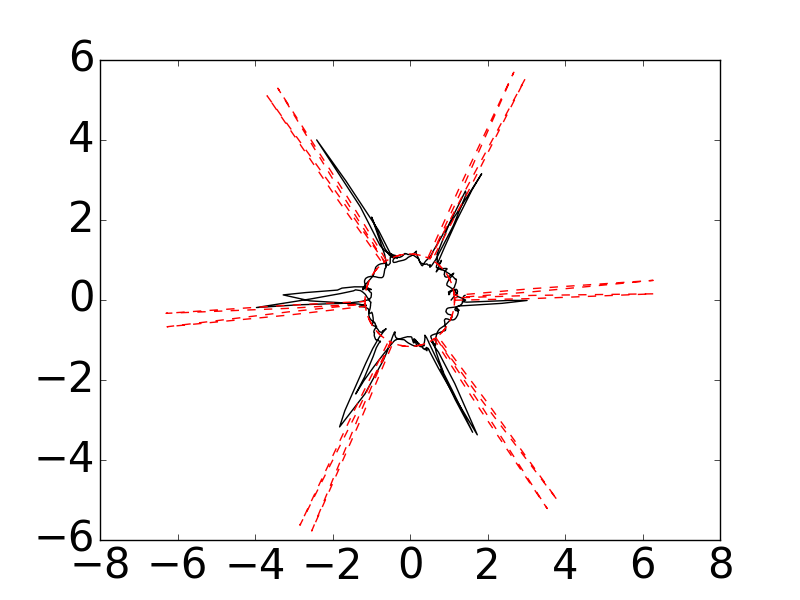}\\
\includegraphics[scale=.2, angle=0]{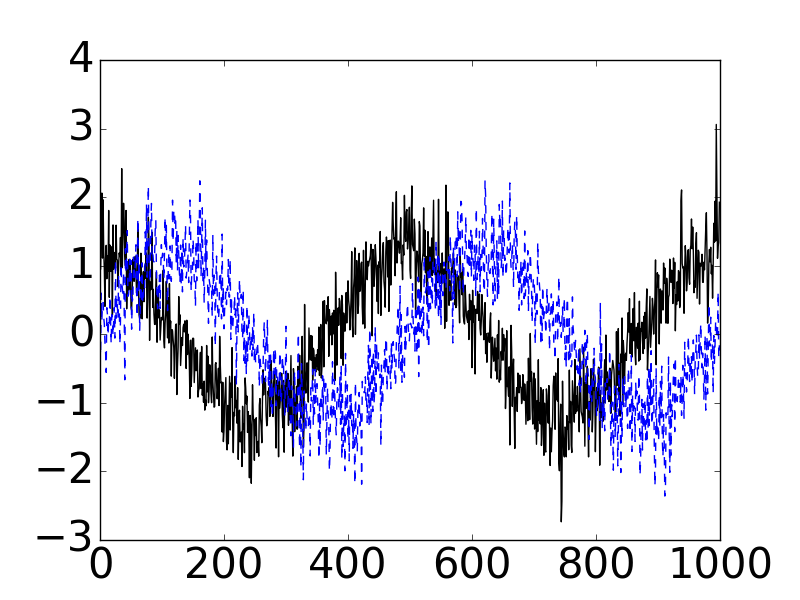} &
\includegraphics[scale=.2, angle=0]{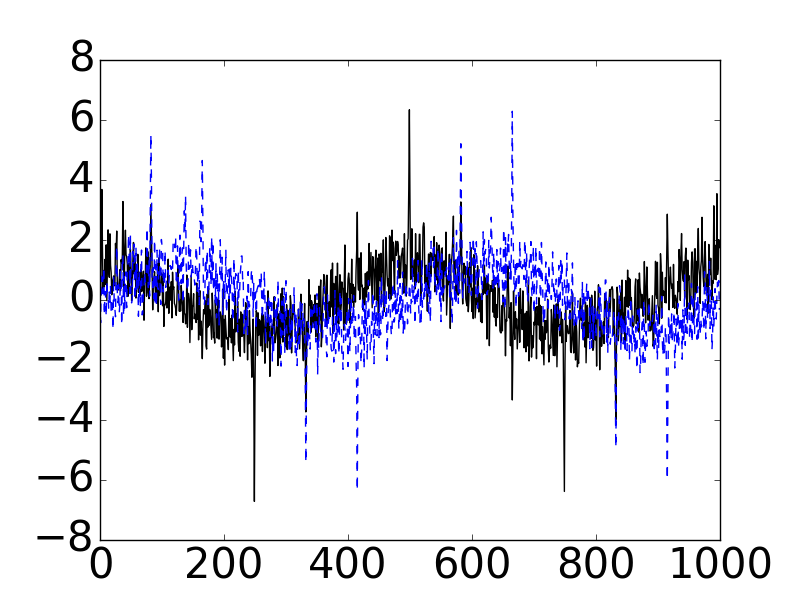} &
\includegraphics[scale=.2, angle=0]{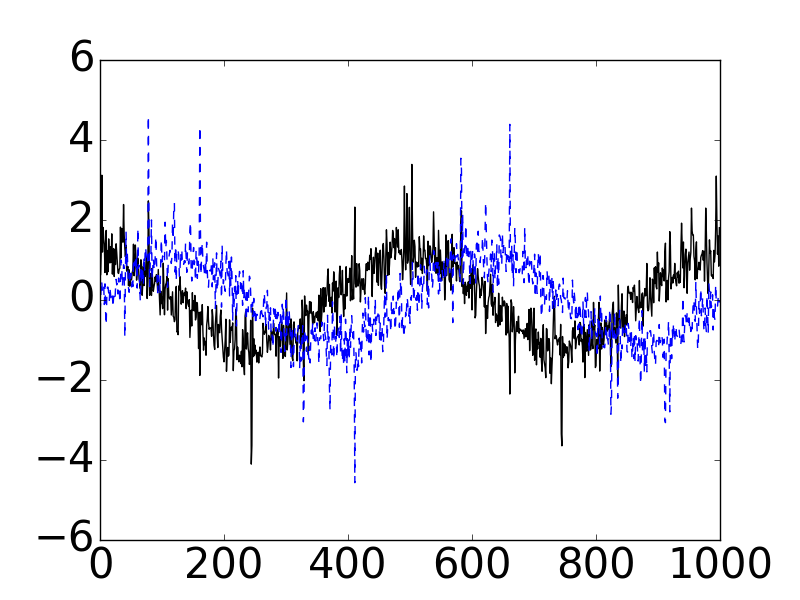} &
\includegraphics[scale=.2, angle=0]{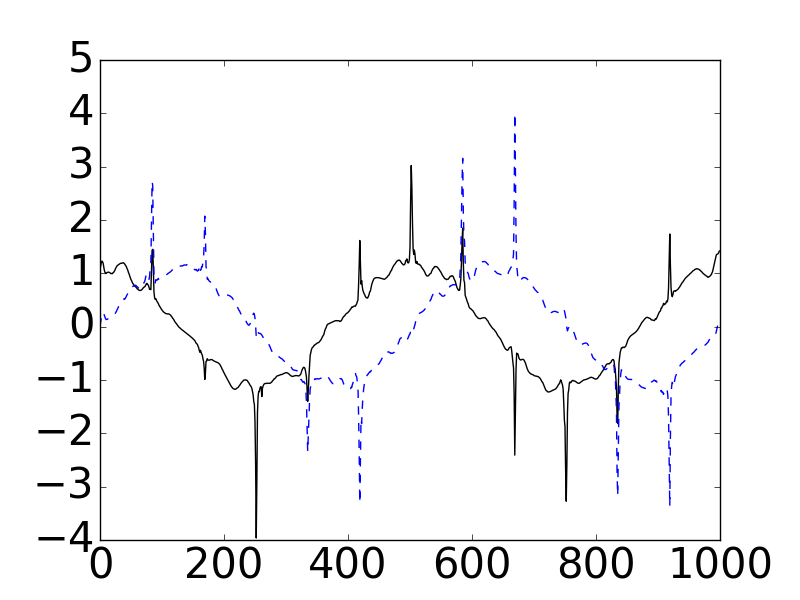}
\end{tabular}
\caption{Centroids obtained from a set of height noisy instances $\{(x_k,y_k)\}_{k=1\cdots 8}$ for Euclidean, DBA, CTW and TEKA averaging   methods. The log power spectra in dB (top), the 2D shape (center) and x,y waveforms (bottom) are shown.} 
\label{fig:noisyCentroids}
\end{figure*}

In practice we have adopted the following setting: $f_0=\omega_o/(2.\pi)=20Hz$, and $A_0= 1$. We then center and normalize this 2D signal to get $(\tilde{X_k}(t),\tilde{Y_k}(t))$ corresponding to the plots given in Figure \ref{fig:cleanShape}. The log power spectrum of the $\tilde{X_k}$ component, that is presented in Figure \ref{fig:cleanSpectra}, shows the Dirac spike located at $f_0=20Hz$ (corresponding to the sine component), and the convolution of this spike with a Dirac comb in the frequency domain that results in pairs of  Dirac spikes symmetrically located ($\pm 20Hz$) around multiples of $6f_0$, namely $120Hz$, $240Hz$, etc.  
This shows that this signal is characterized by an infinite spectrum.

We consider then noise utterances $\epsilon_k(t)$ with zero mean and variance one added to each instances of the 2D signal:
\begin{align}
x_k(t)=\tilde{X_k}(t)+\epsilon_k(t) \nonumber\\
y_k(t)=\tilde{Y_k}(t)+\epsilon_k(t) \nonumber
\end{align}
leading to a signal to noise ratio of $0dB$. An example of such noisy instance is given in Figure \ref{fig:noisyShape}. Because of the scattering of the  random components of the signal in a wide spectral band, traditional noise reduction techniques, such as those presented in \cite{Hassan2010} for instance, will not allow to recover the signal properly.

%\color{red}
The task consists in reducing the noise as far as possible to recover the 2D shape of the noise free signal from a small set of noisy instances $\{(x_k,y_k)\}_{k=1\cdots 8}$ containing two "periods" of the clean signal. Figure \ref{fig:noisyCentroids} presents the centroid shapes obtained using, from left to right, Euclidean, DBA, CTW and TEKA methods respectively. We can see that the Euclidean centroid retrieves partially the low frequency sine component without properly sorting out the spikes components, while DBA more accurately retrieves the spikes, however without achieving to suppress the low frequency noise around the sine component. CTW centroid appears to be in between and achieves partially to reduce the low frequency noise and to extract the spikes. TEKA achieves the best retrieval of the sine and spikes components that are better timely and spatially separated. The spectral analysis presented in Figure \ref{fig:noisyCentroids} (top) gives further insight: for DBA and CTW centroids, top center sub-figures, the series of pairs of Dirac spikes (in dotted red) are still hidden into the noise level (black curve), while it is much more separated from the noise for the TEKA centroid, as shown in the top right side sub-figure.

Moreover, if we take the clean shapes as ground truth, the signal to noise ratio (SNR) gains estimated from the log power spectra (to get rid of the phase) is $0 dB$ for the noisy shapes , while it is $1.58 dB$ for the Euclidean centroid, $1.17 dB$ for the DBA centroid, $1.57 dB$ for the CTW centroid, and $3.88 dB$ for the TEKA centroid. Note that in the calculation of the SNR,  preserving the spikes has a lower impact compared to preserving  the low frequency sine wave, which explains why the SNR values obtained by the DBA and CTW centroid are lower than for the Euclidean centroid.

% spectra:
%    euclid noise reduction gain (db)= 5.56881061738
%    DBA noise reduction gain (db)= 0.962960005553
%    ctw noise reduction gain (db)= 4.58672145318
%    TEKA noise reduction gain (db)= 7.2248995785

%SNR clean TEKA= 0.34812044938
%SNR clean TEKA2= 0.347751589678
%SNR clean DBA= 0.0761715573599
%SNR clean CTW= 0.253333241489
%SNR clean GTW= 0.304528210477
%SNR clean EUCLID= 0.285496072536
%RMSE_NOISY= 64.9181047118   in DB: -0.00281916923553
%RMSE_DBA= 54.1338350749   in DB: 0.0760778673693
%RMSE_CTW= 36.0055602665   in DB: 0.253177089676
%RMSE_GTW= 31.9930214016   in DB: 0.304491406319
%RMSE_TEKA= 28.9415971747   in DB: 0.348024168374
%RMSE_TEKA2= 28.9674946552   in DB: 0.347635727344
%RMSE_EUCLID= 33.4298075467   in DB: 0.285412786456

%Finally it is quite noticeable that a limited number of noisy signal utterances are sufficient to extract the shape of the 2D signal. If we consider only two instances of the noisy signal, the TEKA centroid estimation that is shown in Figure \ref{fig:noisyCentroid2} gives a pretty good idea of the clean shape, although the remaining noise level is a bit higher than when four instances are averaged ($RMSE=0.280$ and $SNR=1.11dB$). This property could find application in stereoscopic sound perception for instance.

In terms of noise reduction, this experiment demonstrates the ability of the TEKA centroid to better recover, from few noisy utterances, a signal whose components are scattered in a wide band spectrum. Indeed, if the noise level increases, the quality of the denoising will be reduced.

\subsection{Discussion}

We believe that the noise filtering ability of TEKA is mainly due to the averaging technique described in the equation (\ref{eq:TEKA}), which aggregates many plausible alignments between samples (instead of a best one) while averaging also the time of occurrence of the samples, in particular those corresponding to expected pattern location and duration such as the CBF shapes or the spike locations in the third experiment. This ability is also likely to explain the best accuracy results obtained by TEKA comparatively to the state of the art methods, CTW and DBA.

Furthermore, it seems that the KDTW measure is more adapted to match centroids than DTW. Here again, handling  several good to best alignments rather than a single optimal one allows for matching the centroids in many ways that are averaged by the measure. This has been verified for CTW in 1-NC classification tasks and is true for TEKA and DBA also. 

The main limitation in exploiting TEKA (and KDTW) is the tuning of the $\nu$ parameter that control the selectivity of the local kernel. $\nu$ is dependent on the length of the time series and need  to be adapted to the task itself. Basically, if $\nu$ is too small TEKA will filter out high frequency events just as a moving average filter. Conversely, if $\nu$ is too high, the computation of the products of local probabilities along the alignment paths will bear some loss of significance in terms of the numerical calculation.
Despite this tuning requirement, the three experiments, that we have carried out in this study, demonstrate its applicability and usefulness.

%\begin{figure}[h!]
%\centering
%\begin{tabular}{ccc}
%\includegraphics[scale=.35, angle=0]{shapeCentroidrealigned2.eps}
%%\includegraphics[scale=.35, angle=0]{noisyShapeCentroidpKDTWpwa2D2.eps}
%\end{tabular}
%\caption{ TEKA centroid obtained from a set of two noisy instances  $\{(x_k,y_k)\}_{k=1\cdots 2}$.}
%\label{fig:noisyCentroid2}
%\end{figure}

\color{black}
\section{Conclusion}
%\color{red}
In this paper, we have addressed the problem of averaging a set of time series in the context of a time elastic distance measure such as Dynamic Time Warping. The new perspective provided by the kernelization of the elastic distance allows a re-interpretation of pairwise kernel alignment matrices as the result of a forward-backward procedure applied on the states of an equivalent stochastic alignment automata. From this re-interpretation, we have proposed a new algorithm, TEKA, based on an iterative agglomerative heuristic method that allows for efficiently computing good solutions to the multi-alignment of time series. This algorithm exhibits quite interesting denoising capabilities which enlarges the area of its potential applications. 

We have presented extensive experiments carried out on synthetic and real data sets, containing univariate but also multivariate time series. Our results show that centroid-based methods significantly outperform medoid-based methods in the context of a first nearest neighbor and SVM classification tasks. More strikingly, the TEKA algorithm, which integrates joint averaging in the sample space and along the time axis, is significantly better than the state-of-the art DBA and CTW algorithms, with a similar algorithmic complexity. It enables robust training set reduction which has been experimented on an isolated gesture recognition task. Finally we have developed a dedicated synthetic test to demonstrate the denoising capability of our algorithm, a property that is not supported at a same level by the other time-elastic centroid methods on this test.  
\color{black}
%The simple iterative heuristic procedure which is used in the TEKA algorithm could be further optimized, using for instance evolutionist approaches that would offer improvement, especially in the context of summarizing large instance sets.

%\begin{figure*}[t!]
%\centering
%\begin{tabular}{ccc}
%    	\includegraphics[scale=.4]{1p1.eps} 
%    	\includegraphics[scale=.4]{1p2.eps}\\
%    	\includegraphics[scale=.4]{mat11.eps}  
%    	\includegraphics[scale=.4]{mat12.eps} 
%\end{tabular}
%\caption{Projections}
%\label{fig:KPCA-GP}
%\end{figure*}

%A₂ - A₁ 7.790620e-06
%A₃ - A₁ 4.411498e-03
%A₄ - A₁ 6.852185e-12
%A₅ - A₁ 2.442491e-15
%A₃ - A₂ 6.210367e-01
%A₄ - A₂ 1.736271e-01
%A₅ - A₂ 9.912380e-03
%A₄ - A₃ 2.617928e-03
%A₅ - A₃ 2.912847e-05
%A₅ - A₄ 8.363461e-01

%\begin{acknowledgements}
 \section*{Acknowledgments}
The authors thank the French Ministry of Research, the Brittany Region and the European Regional Development Fund that partially funded this research. The authors also thank the promoters of the UCR and UCI data repositories for providing the datasets used in this study.
%\end{acknowledgements}
%\newpage

\bibliographystyle{IEEEtran}
\bibliography{biblio}   % name your BibTeX data base

%\begin{thebibliography}{}
%%
%% and use \bibitem to create references. Consult the Instructions
%% for authors for reference list style.
%\end{thebibliography}

%\begin{IEEEbiography}
%[{\includegraphics[width=1in,height=1.25in,clip,keepaspectratio]{pfm2016.png}}]
%{Pierre-Francois Marteau} received his master engineering degree in computer science from Ecole Nationale Sup\'{e}rieure d'Electronique et d'Informatique de Bordeaux in 1984, and his Ph.D. degree in computer science in 1988 from Institut National Polytechnique de Grenoble (Grenoble INP). He experienced a post doctorate position at  University of Geneva in 1989, and at the Institute for Non Linear Sciences at University of California San Diego in 1990. He then worked for eight years as an IT consultant in Bertin Technologies, a private company in Paris, before joining in 1999 the Computer Science Lab. at Universit\'{e} Bretagne Sud where he his a professor. He joins the Institut de Recherche en Informatique et Syst\`{e}mes Al\'{e}atoires (IRISA) in 2012. His current research interests include pattern recognition and machine learning with application in sequential (symbolic and digital) data processing.
%\end{IEEEbiography}

\end{document}